\documentclass[journal]{IEEEtran}
\usepackage{amsmath,graphicx}
\usepackage[T1]{fontenc}
\usepackage{hyperref, times, cite}
\usepackage{url}
\usepackage{enumitem}
\usepackage{mathtools}
\usepackage{bm, bbm}
\usepackage{cool}
\usepackage{amssymb,amsfonts,amsmath,amsthm,amscd,dsfont,mathrsfs}
\usepackage{graphicx,float,psfrag,epsfig,amssymb}
\usepackage{wrapfig}
\usepackage{relsize}
\usepackage{color}
\usepackage{xcolor}
\usepackage{pict2e}
\usepackage{multicol}
\usepackage{algorithm}
\usepackage{algorithmic}
\usepackage{caption}
\usepackage{nameref}
\usepackage{enumerate}
\usepackage{stackrel}

\usepackage{caption}
\usepackage{subcaption}

\usepackage{amsthm}

\begin{document}
\thispagestyle{empty}

\title{Learning Conditional Independence Differential Graphs From Time-Dependent Data}
\author{ Jitendra K.\ Tugnait 
\thanks{J.K.\ Tugnait is with the Department of 
Electrical \& Computer Engineering,
200 Broun Hall, Auburn University, Auburn, AL 36849, USA. 
Email: tugnajk@auburn.edu . }
\thanks{This work was supported by the National Science Foundation under Grant CCF-2308473.}}

\maketitle

\renewcommand{\algorithmicrequire}{\textbf{Input:}}
\renewcommand{\algorithmicensure}{\textbf{Output:}}

\begin{abstract}
Estimation of  differences in conditional independence graphs (CIGs) of two time series Gaussian graphical models (TSGGMs) is investigated where the two TSGGMs are known to have similar structure. The TSGGM structure is encoded in the inverse power spectral density (IPSD) of the time series. In several existing works, one is interested in estimating the difference in two precision matrices to characterize underlying changes in conditional dependencies of two sets of data consisting of independent and identically distributed (i.i.d.) observations. In this paper we consider estimation of the difference in two IPSDs to characterize the underlying changes in conditional dependencies of two sets of time-dependent data. Our approach accounts for data time dependencies unlike past work. We analyze a penalized D-trace loss function approach in the frequency domain for differential graph learning, using Wirtinger calculus. We consider both convex (group lasso) and non-convex (log-sum and SCAD group penalties) penalty/regularization functions. An alternating direction method of multipliers (ADMM) algorithm is presented to optimize the objective function. We establish  sufficient conditions in a high-dimensional setting for consistency (convergence of the inverse power spectral density to true value in the Frobenius norm) and graph recovery. Both synthetic and real data examples are presented in support of the proposed approaches. In synthetic data examples, our log-sum-penalized differential time-series graph estimator significantly outperformed our lasso based differential time-series graph estimator which, in turn, significantly outperformed an existing lasso-penalized i.i.d.\ modeling approach, with $F_1$ score as the performance metric. In a 120-dimensional moving-average model based time series example, for sample sizes of $n=512$ and 4096, our log-sum-penalized estimator improved the $F_1$ scores by 84\% and 44\%, respectively, over our lasso-penalized method and by 119\% and 112\%, respectively, over existing lasso-penalized i.i.d.\ method  ($F_1$ scores 0.46 and 0.91 for log-sum, 0.28 and 0.58 for proposed lasso, and 0.21 and 0.43 for i.i.d.\ lasso).
\end{abstract}


\section{Introduction} \label{intro}
\IEEEPARstart{G}{raphical} models are used to display and explore conditional independence structure in the analysis of multivariate data  \cite{Whittaker1990, Lauritzen1996, Buhlmann2011, Dahlhaus2000}. Consider a graph ${\cal G} = \left( V, {\cal E} \right)$ with a set of $p$ vertices (nodes) $V = \{1,2, \cdots , p\} =[p]$, and a corresponding set of (undirected) edges ${\cal E} \subseteq [p] \times [p]$. Next consider a stationary (real-valued), zero-mean,  $p-$dimensional multivariate Gaussian time series ${\bm x}(t)$, $t=0, \pm 1, \pm 2, \cdots $, with $i$th component $x_i(t)$, and correlation (covariance) matrix function ${\bm R}_{xx}( \tau ) = \mathbb{E} \{ {\bm x}(t + \tau) {\bm x}^T(t ) \}$, $ \tau = 0, \pm 1,  \cdots  $. Given $\{ {\bm x}(t) \}$, in the corresponding graph ${\cal G}$, each component series $\{ x_i(t) \}$ is represented by a node ($i$ in $V$), and associations between components $\{ x_i (t) \}$ and $\{ x_j(t) \}$ are represented by edges between nodes $i$ and $j$ of ${\cal G}$. In a conditional independence graph (CIG), there is no edge between nodes $i$ and $j$ (i.e., $\{ i,j \} \not\in {\cal E}$) if and only if (iff) $x_i(t)$ and $x_j(t)$ are conditionally independent given the remaining $p$-$2$ scalar series $x_\ell(t)$, $\ell \in [p]$, $\ell \neq i$, $\ell \neq j$. (This is a generalization of CIG for random vectors where $\{ i,j \} \not\in {\cal E}$ iff $\Omega_{ij} = 0$ \cite{Whittaker1990, Lauritzen1996, Buhlmann2011}; ${\bm \Omega} = (E\{ {\bm x}(t) {\bm x}^\top(t) \})^{-1}$ is the precision matrix.) 

Let ${\bm S}_x(f)$ denote the power spectral density (PSD) matrix of $\{ {\bm x}(t) \}$, given by ${\bm S}_x(f) = \sum_{\tau = -\infty}^{\infty}  {\bm R}_{xx}( \tau ) e^{-\iota 2 \pi f \tau}$, $\iota=\sqrt{-1}$. In \cite{Dahlhaus2000} it was established that conditional independence of two time series components given all other components of the time series, is encoded by zeros in the inverse PSD (IPSD), that is, $\{ i,j \} \not\in {\cal E}$ iff the $(i,j)$-th element of ${\bm S}_x^{-1}(f)$ vanishes, i.e.,  $[{\bm S}_x^{-1}(f)]_{ij} = 0$ for every $f$. Hence one can use estimated IPSD of observed time series to infer the associated graph. 

There has also been some interest in differential network analysis where one estimates the difference in two inverse covariance matrices \cite{Yuan2017, Jiang2018, Tang2020, Tugnait2024}. Given observations ${\bm x}$ and ${\bm y}$ from two groups of subjects, one is interested in the difference ${\bm \Delta} = {\bm \Omega}_y - {\bm \Omega}_x$, where ${\bm \Omega}_x = (E \{ {\bm x} {\bm x}^\top \} )^{-1}$ and ${\bm \Omega}_y = (E \{ {\bm y} {\bm y}^\top \} )^{-1}$. The associated differential graph is ${\cal G}_\Delta = \left( V, {\cal E}_\Delta \right)$ where $\{i,j\} \in {\cal E}_\Delta$ iff $[{\bm \Delta}]_{ij} \ne 0$. It characterizes differences between the Gaussian graphical models (GGMs) of the two sets of data. We use the term differential graph as in \cite{Zhao2022, Tugnait2024} (\cite{Yuan2017, Tang2020, Wu2020} use the term differential network). As noted in \cite{Yuan2017,Tang2020}, in biostatistics, the differential network/graph describes the changes in conditional dependencies between components under different environmental or genetic conditions. For instance, one may be interested in the differences in the graphical models of healthy and impaired subjects, or models under different disease states, given gene expression data or functional magnetic resonance imaging (fMRI) signals \cite{Danaher2014, Zhao2014, Belilovsky2016}.

In several applications such as fMRI signal analysis or financial time series analysis, the underlying temporal data is not i.i.d. Analysis of some resting state fMRI data in \cite{Shu2019} shows significant time-dependence. The data set analyzed in \cite[Sec.\ 2.2]{Shu2019} consists of a single subject 1190 temporal brain images, each image with 907 functional brain nodes. This data passed stationarity, Gaussianity and linearity tests \cite[Sec.\ 2.2]{Shu2019}, implying that Gaussian time series assumption used in this paper would be appropriate. The focus of \cite{Shu2019} is analysis of graphical models derived from precision matrix of dependent data; they do not address time series graphical models. In \cite[Sec.\ 5.2]{Songsiri2010} autoregressive models are fitted to financial time series (international stock market data) to infer the underlying time series graphical model. Differential network analysis in such applications calls for consideration of time-dependence in the data as well as consideration of time series graphical models, instead of just assuming that the data is i.i.d., as in \cite{Yuan2017, Jiang2018, Tang2020, Wu2020, Tugnait2024}. There is no prior reported work on differential times series graph estimation. This paper attempts to fill this gap.

In this paper we address the problem of estimating differences in two time series Gaussian graphical models (TSGGMs) which are known to have similar structure. Our approach accounts for data time dependencies unlike past work. The TSGGM structure is encoded in its IPSD just as the vector GGM structure is encoded in its precision matrix. We consider estimation of the difference in two IPSD's to characterize underlying changes in conditional dependencies of two sets of time-dependent data $\{ {\bm x}(t) \}_{t=1}^{n_x}$ and $\{ {\bm y}(t) \}_{t=1}^{n_y}$.  We analyze a penalized D-trace loss function approach in the frequency domain for differential graph learning, using Wirtinger calculus \cite{Schreier10}. As a preliminary step, we first address the problem of estimation of complex differential graphs, given two complex-valued i.i.d.\ time series. We consider both convex (group lasso) and non-convex (log-sum and  Smoothly Clipped Absolute Deviation (SCAD) group penalties) penalty/regularization functions. The use of non-convex penalties (unlike convex lasso penalty) is known to yield more accurate results, i.e., they can produce sparse set of solution like lasso, and approximately unbiased coefficients for large coefficients, unlike lasso \cite{Fan2001, Candes2008, Lam2009}. 

\subsection{Related Work}  The problem of estimation of complex differential graphs and the general problem of differential times series graph estimation have not been investigated before.  The work of \cite{Zhao2022} considers time series differential graphs with D-trace loss functions except that in \cite{Zhao2022} ${\bm x}(t)$ and ${\bm y}(t)$ are non-stationary (``functional'' modeling), and instead of a single record (sample) of ${\bm x}(t)$, $t \in [n]$ and ${\bm y}(t)$, $t \in [n]$, as in this paper, they assume multiple independent observations of ${\bm x}(t)$, $t \in {\cal T}$, and ${\bm y}(t)$, $t \in {\cal T}$ (a closed subset of real line). However, as in \cite{Zhao2022}, we follow the framework of \cite{Negahban2012} for theoretical analysis. Unlike our paper, \cite{Zhao2022} does not consider non-convex penalties.

Differential network analysis reported in \cite{Yuan2017, Jiang2018, Tang2020, Wu2020, Tugnait2024} all deal with i.i.d.\ data whereas we address dependent data. In \cite{Na2021} differential latent variable graphical models are estimated assuming i.i.d.\ data. In latent variable models there are some hidden nodes. We do not consider this aspect in this paper. As in \cite{Yuan2017, Jiang2018, Tang2020, Wu2020, Tugnait2024} we use a D-trace loss function approach. One naive approach to differential network analysis with i.i.d.\ data would be to estimate the two precision matrices separately by any existing estimator (see \cite{Buhlmann2011, Lam2009, Zou2008, Tugnait21b} and references therein) and then calculate their difference to estimate the differential graph. In such an approach one estimates twice the number of parameters (the respective precision matrices instead of the difference), therefore, one needs larger sample sizes for same accuracy. Also, this naive approach imposes sparsity constraints on each precision matrix for the methods to work. The same comment applies to methods such as \cite{Danaher2014}, where the two precision matrices and their differences are jointly estimated. If only the difference in the precision matrices is of interest,  direct estimation of the difference in the two precision matrices is preferable and has been considered in \cite{Zhao2014, Yuan2017, Jiang2018, Tang2020, Wu2020, Tugnait2024}, where only the difference is required to be sparse, not the two individual precision matrices. In \cite{Yuan2017, Jiang2018, Tang2020, Wu2020, Tugnait2024} precision difference matrix estimators are based on  a D-trace loss \cite{Zhang2014}, while \cite{Zhao2014} discusses a Dantzig selector type estimator. In this paper we extend the D-trace loss framework of \cite{Yuan2017, Jiang2018} for i.i.d.\ data to address time-dependent data via a frequency-domain formulation.

Our work exploits prior work on graphical modeling of real time series in high-dimensional settings. Nonparametric approaches for graphical modeling of real time series in high-dimensional settings ($p$ is large and/or sample size $n$ is of the order of $p$) have been investigated  in \cite{Jung2015a, Tugnait2022, Tugnait2025c}, among others. We use the frequency-domain formulation of \cite{Tugnait2022, Tugnait2025c} which deals with graphical modeling of time series, but not with differential graphical modeling addressed here. In \cite{Krampe2025, Chang2025} estimation of high-dimensional power spectral matrix is addressed. Time series graphical modeling is not discussed in \cite{Chang2025}, unlike \cite{Krampe2025} where testing-based methods are used for inference of graphical models (as opposed to regularization based methods in  \cite{Jung2015a, Tugnait2022, Tugnait2025c}). Differential graphical modeling is not addressed in \cite{Krampe2025, Chang2025}. 

Although non-convex penalties have been extensively used for graph estimation (see \cite{Lam2009, Zou2008, Tugnait21b} and references therein) and recently for differential graph estimation from i.i.d.\ data \cite{Tugnait2025a}, they have not been applied to differential time-series graphs.  For optimization we use the standard alternating direction method of multipliers (ADMM) approaches \cite{Boyd2010} except that our ADMM algorithm applies to real objective function of complex variables exploiting the Wirtinger calculus. Our numerical results show that our log-sum-penalized differential time-series graph estimator significantly outperforms our lasso based differential time-series graph estimator which in turn, significantly outperforms  the i.i.d.\ modeling based time domain methods of \cite{Yuan2017, Jiang2018} (lasso penalty) and \cite{Tugnait2025a} (log-sum penalty), with $F_1$ score as the performance metric.

Graphical models have also been inferred from consideration other than statistical \cite{Dong2019}. One class of graphical models are based on signal smoothness \cite{Dong2016, Dong2019, Kalofolias2019} where graph learning from data becomes equivalent to estimation of the graph Laplacian matrix. Some reviews of various graph learning approaches may be found in \cite{Qiao2018}, \cite{Xia2021}, \cite{Xia2025} and \cite{Chen2024}. A large variety of graph learning models and approaches exist, motivated by diverse applications in signal processing, machine learning, and other areas. In \cite{Xia2025} (also \cite{Xia2021}) existing graph learning methods are classified into four broad categories: deep learning based methods, matrix factorization based methods, random walk based methods, and graph signal processing based methods. In terms of these four categories, our approach falls in the category of graph signal processing based methods with the sub-category of ``learning topology structure.'' Differently from \cite{Xia2025}, \cite{Qiao2018} categorizes graph learning methods based on two graph construction steps: (1) determine the edge set ${\cal E}$, called $E$-step, and (2) based on ${\cal E}$, determine an edge weight matrix ${\bm W}$, called $W$-step, even though in some methods these two steps may be merged into one, or the second step may be executed first yielding ${\bm W}$ which then determines ${\cal E}$. In \cite{Chen2024} graphical modeling is approached from a statistical viewpoint and a wide variety of models (i.i.d.\ Gaussian data, matrix-valued data, quantile graphical models, etc.) and approaches are considered. In this paper we are interested in conditional independence differential graphs, a topic not addressed in \cite{Qiao2018}, \cite{Xia2021}, \cite{Xia2025} and \cite{Chen2024}.

More recently there has been interest in introducing fairness considerations in graphical modeling \cite{Zhou2024} (based on i.i.d.\ data assumption), and in exploiting transfer learning ideas in estimating one target graph and several auxiliary graphs using (i.i.d.) data from multiple sources \cite{Zhang2025}. In \cite{Zhang2025} concepts similar to differential graphs are used in transferring auxiliary graph structure to the target graph.

A class of graph and graph-based learning approaches are motivated by specific application tasks such as clustering and  classification. Examples of such approaches include \cite{Kang2021, Amjad2024, Amjad2025} and relevant references in  \cite{Qiao2018, Xia2021, Xia2025, Chen2024}. While these approaches address important useful problems, they are not related to the differential time series graph learning problem addressed in this paper. In such approaches an important consideration is how to incorporate prior information relevant to the intended application, in the graph model. For instance, both local and global structure information is incorporated in the model of \cite{Kang2021}, together with a rank constraint on the graph Laplacian to reflect the number of clusters. In \cite{Amjad2025} a multi-domain speech emotion recognition problem is addressed where domain discrepancy between target and source domains is captured by similarity and dissimilarity graphs, modeled via Laplacian matrices. Construction of these Laplacian matrices in \cite{Amjad2025} does not follow any statistical approach or consider conditional independence. That is, the objectives in this paper and \cite{Amjad2025} are quite different, resulting in distinctly different approaches. In \cite{Amjad2024} the speech emotion detection problem of \cite{Amjad2025} is combined with gender prediction, and the focus is on optimization of feature selection. In \cite{Amjad2024} a particle swarm optimization approach is investigated which is a general heuristic approach which does not guarantee a global optimum. In this paper our penalized D-trace loss function with lasso penalty or local-linear approximated non-convex penalty, is convex and our ADMM optimization algorithm is provably convergent to a global minimum (see Secs.\ \ref{conv1} and \ref{conv2}).  For such problems ADMM is generally considered to be a computationally  efficient  optimizer \cite{Boyd2010}.
 
As noted in \cite{Qiao2018}, ``... how to select a suitable graph construction/learning strategy in practice ... is a challenging problem without a universal solution, since it depends on many factors ...''

\subsection{Our Contributions}
We first address the problem of estimation of complex differential graphs, given two complex-valued i.i.d.\ time series. These results form the basis for analyzing a novel penalized D-trace loss function approach in the frequency domain for differential graph learning, using Wirtinger calculus. We consider both convex group lasso and non-convex (log-sum and SCAD) group penalties regularization functions. An ADMM algorithm is presented to optimize the objective function, using a local linear approximation (LLA) \cite{Zou2008, Lam2009} based  iterative approach for non-convex penalties. Theoretical analysis establishing sufficient conditions for consistency (convergence of the inverse power spectral density to true value in the Frobenius norm) and graph recovery is presented using the framework of \cite{Negahban2012} which does not apply to the SCAD penalty. Both synthetic and real data examples are presented in support of the proposed approaches.

A preliminary version of this paper appears in a conference paper \cite{Tugnait2023b} where non-convex penalties are not considered and no proof is given for \cite[Theorem 1]{Tugnait2023b} (corresponding to our Theorem 1). Moreover, \cite{Tugnait2023b} has no counterparts to our Sec.\ \ref{BIC} and Theorem 2, and it has limited synthetic data results and no real data results.

\subsection{Notation and Outline} \label{outnot}
For a set $V$, $|V|$ denotes its cardinality. Given ${\bm A} \in \mathbb{C}^{p \times p}$, we use $\phi_{\min }({\bm A})$, $\phi_{\max }({\bm A})$, $|{\bm A}|$ and $\mbox{tr}({\bm A})$ to denote the minimum eigenvalue, maximum eigenvalue, determinant and  trace of ${\bm A}$, respectively, and we use ${\bm A} \succ {\bm 0}$ and ${\bm A} \succeq {\bm 0}$ to denote that ${\bm A}$ is positive-definite and positive semi-definite, respectively. Given ${\bm B} \in \mathbb{C}^{p \times m}$, $[{\bm B}]_{ij}$ denotes the $(i,j)$-th element of ${\bm B}$, and so does $B_{ij}$, and  ${\bm I}_q$ denotes the $q \times q$ identity matrix. The symbol $\otimes$ denotes the matrix Kronecker product and the symbol $\circ$ denotes the Hadamard product. The superscripts $\ast$ and $H$ denote the complex conjugate and the Hermitian (conjugate transpose) operations, respectively. 

For ${\bm B} \in \mathbb{C}^{p \times q}$, we define the operator norm, the Frobenius norm and the vectorized $\ell_1$ norm, respectively, as $\|{\bm B}\| = \sqrt{\phi_{\max }({\bm B}^H  {\bm B})}$, $\|{\bm B}\|_F = \sqrt{\mbox{tr}({\bm B}^H  {\bm B})}$, $\|{\bm B}\|_1 = \sum_{i,j} |B_{ij}|$ and $\|{\bm B}\|_\infty = \max_{i,j} |B_{ij}|$. For vector ${\bm \theta} \in \mathbb{C}^p$, we define $\| {\bm \theta} \|_1 = \sum_{i=1}^p |\theta_i|$ and $\| {\bm \theta} \|_2 = \sqrt{\sum_{i=1}^p |\theta_i|^2}$, and we also use $\| {\bm \theta} \|$ for $\| {\bm \theta} \|_2$. Given ${\bm A} \in \mathbb{C}^{n \times p}$, column vector $\mbox{vec}({\bm A}) \in \mathbb{C}^{np}$ denotes the vectorization of ${\bm A}$ which stacks the columns of the matrix ${\bm A}$, and ${\rm Re}({\bm A})$ and ${\rm Im}({\bm A})$ denote the real and imaginary parts, respectively, of ${\bm A}$. The notation ${\bm x} \sim {\mathcal N}_c( {\bm m}, \bm{\Sigma})$ denotes a random vector  ${\bm x}$ that is circularly symmetric (proper) complex Gaussian with mean ${\bm m}$  and covariance $\bm{\Sigma}$. Similarly, ${\bm x} \sim {\mathcal N}_r( {\bm m}, \bm{\Sigma})$ denotes a random vector  ${\bm x}$ that is real-valued Gaussian with mean ${\bm m}$  and covariance $\bm{\Sigma}$. Given a variable vector ${\bm x}$ or matrix ${\bm X}$, we use ${\bm x}^\diamond$ or ${\bm X}^\diamond$, respectively, to denote their true values.

The rest of the paper is organized as follows. 
A penalized D-trace loss function is presented in Sec.\ \ref{CDG} for estimation of complex differential graphs, given two complex-valued i.i.d.\ time series. These results form the basis for a novel penalized D-trace loss function approach in the frequency domain for differential graph learning, formulated in Secs.\ \ref{SM} and \ref{PDTL}. A solution to optimization of the penalized D-trace loss is provided in Sec.\ \ref{OPT} and the selection of the tuning parameters is presented in Sec.\ \ref{BIC}. In Sec.\ \ref{TA} we provide a theoretical analysis of the proposed approach, resulting in Theorems 1 and 2. Numerical results are presented in Secs.\ \ref{NE} and \ref{NEreal}. A derivation of (\ref{eqn120}) and the proofs of Theorems 1 and 2 are given in the two appendices.

\section{Complex Differential Graphs} \label{CDG}
As a preliminary step, we first address the problem of estimation of sparse complex differential graphs, given two complex-valued i.i.d.\ time series. To this end, we first review the problem of estimation of real differential graphs in Sec.\ \ref{RGV}. The results of Sec.\ \ref{RGV} are then extended to complex  differential graphs in Secs.\ \ref{PCGV} and \ref{OPCGV}. The results of Secs.\ \ref{PCGV} and \ref{OPCGV} are later exploited in Secs. \ref{SM}, \ref{PDTL} and \ref{OPT} to address time series differential graphs.

\subsection{Real Gaussian Vectors} \label{RGV}
We first recall a formulation of \cite{Yuan2017, Jiang2018, Tang2020, Wu2020} for real-valued data. 
Let ${\bm x} \in \mathbb{R}^{p}$, ${\bm x} \sim {\mathcal N}_r( {\bm 0}, \bm{\Sigma}_{x}^\diamond)$, $ \bm{\Sigma}_{x}^\diamond \succ {\bm 0}$, and suppose we are given i.i.d.\ samples $\{ {\bm x}(t) \}_{t=1}^{n_x}$ of ${\bm x}$, and similarly given i.i.d.\ samples $\{ {\bm y}(t) \}_{t=1}^{n_y}$ of independent ${\bm y} \in \mathbb{R}^{p}$, ${\bm y} \sim {\mathcal N}_r( {\bm 0}, \bm{\Sigma}_{y}^\diamond)$, $ \bm{\Sigma}_{y}^\diamond \succ {\bm 0}$. Let ${\bm \Omega}_{y}^\diamond = (\bm{\Sigma}_{y}^\diamond)^{-1}$ and ${\bm \Omega}_{x}^\diamond = (\bm{\Sigma}_{x}^\diamond)^{-1}$ denote the respective precision matrices, and let $\hat{\bm \Sigma}_x = \frac{1}{n_x} \sum_{t=1}^{n_x} {\bm x}(t) {\bm x}^\top(t)$ and  $\hat{\bm \Sigma}_y = \frac{1}{n_y} \sum_{t=1}^{n_y} {\bm y}(t) {\bm y}^\top(t)$ denote the sample covariance estimates. In \cite{Yuan2017, Jiang2018, Tang2020, Wu2020} one seeks to estimate ${\bm \Delta}^\diamond = {\bm \Omega}_{y}^\diamond - {\bm \Omega}_{x}^\diamond$ and graph ${\cal G}_\Delta = \left( V, {\cal E}_\Delta \right)$, based on $\hat{\bm \Sigma}_x$ and $\hat{\bm \Sigma}_y$.

In \cite{Yuan2017} (see also \cite[Sec.\ 2.1]{Jiang2018}), the following convex D-trace loss function is used for ${\bm \Delta} \in \mathbb{R}^{p \times p}$
\begin{equation}
  L_r({\bm \Delta}, \hat{\bm \Sigma}_x , \hat{\bm \Sigma}_y) = \frac{1}{2} \mbox{tr} (\hat{\bm \Sigma}_x {\bm \Delta} \hat{\bm \Sigma}_y {\bm \Delta}^\top) 
	   - \mbox{tr} ({\bm \Delta}  (\hat{\bm \Sigma}_x-\hat{\bm \Sigma}_y)) \label{eqn15}
\end{equation}
where D-trace refers to difference-in-trace loss function, a term coined in \cite{Zhang2014} in the context of graphical model estimation. 

Using the rule \cite[p.\ 13, (117)]{Petersen2012} regarding matrix differentiation of a trace function, we have
\begin{align}
  \frac{\partial \mbox{tr} (\hat{\bm \Sigma}_x {\bm \Delta} \hat{\bm \Sigma}_y {\bm \Delta}^\top)}{\partial {\bm \Delta}}
	  & = 2 \hat{\bm \Sigma}_x {\bm \Delta} \hat{\bm \Sigma}_y  \label{0eqn15}
\end{align}
and  using \cite[p.\ 12, (100)]{Petersen2012}, we have
\begin{align}
  \frac{\partial \mbox{tr} ({\bm \Delta}  (\hat{\bm \Sigma}_x-\hat{\bm \Sigma}_y))}{\partial {\bm \Delta}}
	  & =  \hat{\bm \Sigma}_x - \hat{\bm \Sigma}_y \, .  \label{1eqn15}
\end{align}
It then follows that
\begin{align}   \label{2eqn15}
  \frac{\partial L_r({\bm \Delta}, \hat{\bm \Sigma}_x , \hat{\bm \Sigma}_y)}{\partial {\bm \Delta}}
	  & = \hat{\bm \Sigma}_x {\bm \Delta} \hat{\bm \Sigma}_y - (\hat{\bm \Sigma}_x - \hat{\bm \Sigma}_y)
\end{align}
and therefore, at the true covariances, we have 
\begin{align}   \label{3eqn15}
  \frac{\partial L_r({\bm \Delta}, \bm{\Sigma}_{x}^\diamond , \bm{\Sigma}_{y}^\diamond)}{\partial {\bm \Delta}}
	  & = \bm{\Sigma}_{x}^\diamond {\bm \Delta} \bm{\Sigma}_{y}^\diamond 
		 - (\bm{\Sigma}_{x}^\diamond - \bm{\Sigma}_{y}^\diamond) \, .
\end{align}
When ${\bm \Delta} = {\bm \Delta}^\diamond = {\bm \Omega}_{y}^\diamond - {\bm \Omega}_{x}^\diamond$, we have
\begin{align}   
   &  \bm{\Sigma}_{x}^\diamond {\bm \Delta} \bm{\Sigma}_{y}^\diamond 
		  =	\bm{\Sigma}_{x}^\diamond ({\bm \Omega}_{y}^\diamond 
			 - {\bm \Omega}_{x}^\diamond) \bm{\Sigma}_{y}^\diamond \nonumber \\
	& \quad = \bm{\Sigma}_{x}^\diamond {\bm \Omega}_{y}^\diamond \bm{\Sigma}_{y}^\diamond
	  - \bm{\Sigma}_{x}^\diamond {\bm \Omega}_{x}^\diamond \bm{\Sigma}_{y}^\diamond
		  = \bm{\Sigma}_{x}^\diamond -  \bm{\Sigma}_{y}^\diamond \, ,
			   \label{4eqn15}
\end{align}
and therefore, by (\ref{3eqn15}), 
\begin{align}   \label{5eqn15}
  \frac{\partial L_r({\bm \Delta}, \bm{\Sigma}_{x}^\diamond , \bm{\Sigma}_{y}^\diamond)}{\partial {\bm \Delta}} 
	   \, \Big|_{{\bm \Delta} = {\bm \Delta}^\diamond}
	  & = {\bm 0} \, .
\end{align}

Using $\mbox{tr}({\bm A}^\top {\bm B} {\bm C} {\bm D}^\top) = \mbox{vec}(\bm A)^\top ({\bm D} \otimes {\bm B}) \mbox{vec}(\bm C)$ and letting ${\bm d} := \mbox{vec}({\bm \Delta}) \in \mathbb{R}^{p^2}$, we have
\begin{align}   
  L_r({\bm \Delta}, \bm{\Sigma}_{x}^\diamond , \bm{\Sigma}_{y}^\diamond)
	  & = \frac{1}{2} {\bm d}^\top (\bm{\Sigma}_{y}^\diamond \otimes \bm{\Sigma}_{x}^\diamond) {\bm d} 
		 - {\bm d}^\top \mbox{vec}(\bm{\Sigma}_{x}^\diamond -  \bm{\Sigma}_{y}^\diamond) \, . \label{6eqn15}
\end{align}
Thus $L_r({\bm \Delta}, \bm{\Sigma}_{x}^\diamond , \bm{\Sigma}_{y}^\diamond)$ is quadratic in ${\bm d}$ with the Hessian matrix given by 
\begin{align}   
  {\bm H}_r
	  & = \bm{\Sigma}_{y}^\diamond \otimes \bm{\Sigma}_{x}^\diamond \, ,
		  \quad [{\bm H}_r]_{k \ell} = 
			 \frac{\partial^2 L_r({\bm \Delta}, \bm{\Sigma}_{x}^\diamond , \bm{\Sigma}_{y}^\diamond)}
			 {\partial [\bm d]_k \partial [\bm d]_\ell } \, . \label{7eqn15}
\end{align}
The eigenvalues of ${\bm H}_r$ are the product of the eigenvalues of $\bm{\Sigma}_{y}^\diamond$ and $\bm{\Sigma}_{x}^\diamond$. By assumption $\bm{\Sigma}_{y}^\diamond$ and $\bm{\Sigma}_{x}^\diamond$ are positive-definite, hence, ${\bm H}_r \succ {\bm 0}$ since all eigenvalues of Hermitian ${\bm H}_r$ are positive. Therefore, the  function $L_r({\bm \Delta}, \bm{\Sigma}_{x}^\diamond , \bm{\Sigma}_{y}^\diamond)$ is strictly convex in ${\bm \Delta}$ and by (\ref{5eqn15}), has a unique minimum at ${\bm \Delta}^\diamond = {\bm \Omega}_{y}^\diamond - {\bm \Omega}_{x}^\diamond$ \cite{Yuan2017, Jiang2018}. 

Since the true data covariances are unavailable,  one uses sample covariances of the data and ${\bm \Delta}$ is estimated by minimizing a lasso-penalized D-trace loss function (\ref{eqn15}) \cite{Yuan2017, Jiang2018, Tang2020, Wu2020}, given by $L_r({\bm \Delta}, \hat{\bm \Sigma}_x , \hat{\bm \Sigma}_y) + \lambda \,  \sum_{k,\ell=1}^p |\Delta_{k \ell}|$. With $V = [p]$ and ${\cal E} \subseteq [p] \times [p]$ the associated differential graph is ${\cal G}_\Delta = \left( V, {\cal E}_\Delta \right)$ where $\{i,j\} \in {\cal E}_\Delta$ iff $[{\bm \Delta}]_{ij} \ne 0$.

\subsection{Proper Complex Gaussian Vectors} \label{PCGV}
Consider complex-valued ${\bm x}, \, {\bm y} \in \mathbb{C}^{p}$, with ${\bm x} \sim {\cal N}_c({\bm 0}, {\bm \Sigma}_{x}^\diamond)$ and ${\bm y} \sim {\cal N}_c({\bm 0}, {\bm \Sigma}_{y}^\diamond)$ with ${\bm \Sigma}_{x}^\diamond \succ {\bm 0}$ and ${\bm \Sigma}_{y}^\diamond \succ {\bm 0}$. Given i.i.d.\ measurements $\{ {\bm x}(t) \}_{t=1}^{n_x}$ and $\{ {\bm y}(t) \}_{t=1}^{n_y}$, we desire to estimate ${\bm \Delta}^\diamond = {\bm \Omega}_{y}^\diamond - {\bm \Omega}_{x}^\diamond$. 

We propose a real-valued cost of complex-valued ${\bm \Delta} \in \mathbb{C}^{p \times p}$ as
\begin{align}
 & L({\bm \Delta}, \hat{\bm \Sigma}_x , \hat{\bm \Sigma}_y) = 
	 \frac{1}{2} \Big( \mbox{tr} (\hat{\bm \Sigma}_x {\bm \Delta} \hat{\bm \Sigma}_y {\bm \Delta}^H) 
	 + \mbox{tr} (\hat{\bm \Sigma}_x^\ast {\bm \Delta}^\ast \hat{\bm \Sigma}_y^\ast {\bm \Delta}^\top) \Big)  \nonumber \\
	& \quad   - \mbox{tr} \Big({\bm \Delta}  (\hat{\bm \Sigma}_x-\hat{\bm \Sigma}_y) + 
		 {\bm \Delta}^\ast (\hat{\bm \Sigma}_x^\ast-\hat{\bm \Sigma}_y^\ast) \Big) \label{deqn200}
\end{align}
with $\hat{\bm \Sigma}_x = \frac{1}{n_x} \sum_{t=1}^{n_x} {\bm x}(t) {\bm x}^H(t)$ and $\hat{\bm \Sigma}_y = \frac{1}{n_y} \sum_{t=1}^{n_y} {\bm y}(t) {\bm y}^H(t)$.

We will use Wirtinger calculus \cite[Appendix 2]{Schreier10} to analyze $L({\bm \Delta}, \hat{\bm \Sigma}_x , \hat{\bm \Sigma}_y)$ where we view it as a real-valued function of two ``independent'' complex-valued vectors $\mbox{vec}({\bm \Delta})$ and its conjugate $\mbox{vec}({\bm \Delta}^\ast)$. Similar to  (\ref{0eqn15}) but using \cite[p.\ 12, (100),(103)]{Petersen2012} and the fact that $\hat{\bm \Sigma}_x$ and $\hat{\bm \Sigma}_y$ are Hermitian, we have 
\begin{align}
  \frac{\partial \mbox{tr} (\hat{\bm \Sigma}_x {\bm \Delta} \hat{\bm \Sigma}_y {\bm \Delta}^H)}{\partial {\bm \Delta}^\ast}
	  & =  \hat{\bm \Sigma}_x {\bm \Delta} \hat{\bm \Sigma}_y  
		= \frac{\partial \mbox{tr} (\hat{\bm \Sigma}_x^\ast {\bm \Delta}^\ast \hat{\bm \Sigma}_y^\ast {\bm \Delta}^\top)}
		      {\partial {\bm \Delta}^\ast}  \label{10eqn15}
\end{align}
and similar to  (\ref{1eqn15}), we have
\begin{align}
  \frac{\partial \mbox{tr} ({\bm \Delta}^\ast (\hat{\bm \Sigma}_x^\ast-\hat{\bm \Sigma}_y^\ast))}
	   {\partial {\bm \Delta}^\ast}
	  & =  \hat{\bm \Sigma}_x - \hat{\bm \Sigma}_y \, . \label{11eqn15}
\end{align}
Note that  $\frac{\partial \mbox{tr} ({\bm \Delta} (\hat{\bm \Sigma}_x-\hat{\bm \Sigma}_y))}
	   {\partial {\bm \Delta}^\ast} = {\bm 0}$. It then follows that
\begin{align}   \label{12eqn15}
  \frac{\partial L({\bm \Delta}, \hat{\bm \Sigma}_x , \hat{\bm \Sigma}_y)}{\partial {\bm \Delta}^\ast}
	  & = \hat{\bm \Sigma}_x {\bm \Delta} \hat{\bm \Sigma}_y - (\hat{\bm \Sigma}_x - \hat{\bm \Sigma}_y)
\end{align}
and therefore, at the true covariances, we have 
\begin{align}   \label{13eqn15}
  \frac{\partial L({\bm \Delta}, \bm{\Sigma}_{x}^\diamond , \bm{\Sigma}_{y}^\diamond)}{\partial {\bm \Delta}^\ast}
	  & = \bm{\Sigma}_{x}^\diamond {\bm \Delta} \bm{\Sigma}_{y}^\diamond 
		 - (\bm{\Sigma}_{x}^\diamond - \bm{\Sigma}_{y}^\diamond) \, .
\end{align}
When ${\bm \Delta} = {\bm \Delta}^\diamond = {\bm \Omega}_{y}^\diamond - {\bm \Omega}_{x}^\diamond$, by (\ref{4eqn15}), we have 
\begin{align}   \label{15eqn15}
  \frac{\partial L({\bm \Delta}, \bm{\Sigma}_{x}^\diamond , \bm{\Sigma}_{y}^\diamond)}{\partial {\bm \Delta}^\ast} 
	   \, \Big|_{{\bm \Delta} = {\bm \Delta}^\diamond}
	  & = {\bm 0} \, .
\end{align}

Define 
\begin{align}
  {\bm \theta}&  = [\mbox{vec}({\bm \Delta})^\top \; \mbox{vec}({\bm \Delta})^H]^\top \, \in \, \mathbb{C}^{2 p^2} \, , \\
   {\bm H} & = \begin{bmatrix} ({\bm \Sigma}_{y}^\diamond)^\ast \otimes {\bm \Sigma}_{x}^\diamond  & {\bm 0} \\
	                 {\bm 0} & {\bm \Sigma}_{y}^\diamond \otimes ({\bm \Sigma}_{x}^\diamond)^\ast \end{bmatrix} \, , 
						\label{17eqn15} \\
		{\bm b} & = \begin{bmatrix} \mbox{vec}({\bm \Sigma}_{x}^\diamond-{\bm \Sigma}_{y}^\diamond) \\ 
					 \mbox{vec}(({\bm \Sigma}_{x}^\diamond)^\ast-({\bm \Sigma}_{y}^\diamond)^\ast) \end{bmatrix} \, .
\end{align}
Using $\mbox{tr}({\bm A}^\top {\bm B} {\bm C} {\bm D}^\top) = \mbox{vec}(\bm A)^\top ({\bm D} \otimes {\bm B}) \mbox{vec}(\bm C)$, we have 
\begin{equation}
L({\bm \Delta}, {\bm \Sigma}_{x}^\diamond , {\bm \Sigma}_{y}^\diamond) 
 = \frac{1}{2} {\bm \theta}^H {\bm H} {\bm \theta} - 
           {\bm \theta}^H {\bm b} \, .
\end{equation}
Clearly $L({\bm \Delta}, {\bm \Sigma}_{x}^\diamond , {\bm \Sigma}_{y}^\diamond)$ is quadratic in ${\bm \theta}$ with the complex augmented Hessian matrix \cite[Def.\ A2.5]{Schreier10} ${\bm H}$ given by (\ref{17eqn15}), satisfying
\begin{align}   
   [{\bm H}]_{k \ell} = 
			 \frac{\partial^2 L({\bm \Delta}, {\bm \Sigma}_{x}^\diamond , {\bm \Sigma}_{y}^\diamond)}
			  {\partial [\bm \theta]_k \partial [\bm \theta]_\ell^\ast } \, . \label{18eqn15}
\end{align}
The eigenvalues of ${\bm H}$ are the product of the eigenvalues of $\bm{\Sigma}_{y}^\diamond$ and $\bm{\Sigma}_{x}^\diamond$. By assumption $\bm{\Sigma}_{y}^\diamond$ and $\bm{\Sigma}_{x}^\diamond$ are Hermitian positive-definite, hence, ${\bm H} \succ {\bm 0}$ since all eigenvalues of Hermitian ${\bm H}$ are positive. Therefore, the  function $L({\bm \Delta}, {\bm \Sigma}_{x}^\diamond , {\bm \Sigma}_{y}^\diamond)$ is strictly convex in ${\bm \Delta}$ and by (\ref{15eqn15}), has a unique minimum at ${\bm \Delta}^\diamond = {\bm \Omega}_{y}^\diamond - {\bm \Omega}_{x}^\diamond$.

Since the true data covariances are unavailable,  one uses sample covariances of the data as in (\ref{deqn200}).  In this case we replace ${\bm \Sigma}_{x}^\diamond$ and ${\bm \Sigma}_{y}^\diamond$ in (\ref{17eqn15}) with $\hat{\bm \Sigma}_x$ and $\hat{\bm \Sigma}_y$, respectively, resulting in the Hessian matrix $\hat{\bm H}$ 
\begin{align}
   \hat{\bm H} & = \begin{bmatrix} \hat{\bm \Sigma}_y^\ast \otimes \hat{\bm \Sigma}_x  & {\bm 0} \\
	                 {\bm 0} & \hat{\bm \Sigma}_y \otimes \hat{\bm \Sigma}_x^\ast \end{bmatrix} \, .
						\label{117eqn15}
\end{align} 
Since $\hat{\bm \Sigma}_x \succeq {\bm 0}$ and $\hat{\bm \Sigma}_y \succeq {\bm 0}$, we now have $\hat{\bm H} \succeq {\bm 0}$, implying that $L({\bm \Delta}, \hat{\bm \Sigma}_x , \hat{\bm \Sigma}_y)$ is convex, but not necessarily strictly convex.  In the high-dimensional case of $\min\{n_x,n_y \}$ less than or of the order of $p$, to enforce sparsity and to make the problem well-conditioned, for $\lambda > 0$, define the lasso-penalized D-trace loss (similar to \cite{Yuan2017, Jiang2018, Tang2020, Wu2020})
\begin{align}
  & L_\lambda({\bm \Delta}, \hat{\bm \Sigma}_x , \hat{\bm \Sigma}_y) =
	  L({\bm \Delta}, \hat{\bm \Sigma}_x , \hat{\bm \Sigma}_y) + \lambda \,  \sum_{k,\ell=1}^p |\Delta_{k \ell}| \, .
		    \label{eqn25}
\end{align}
We seek $\hat{\bm \Delta} = \arg\min_{\bm \Delta} L_\lambda({\bm \Delta}, \hat{\bm \Sigma}_x , \hat{\bm \Sigma}_y)$. We discuss this aspect next in Sec.\ \ref{OPCGV}. With $V = [p]$ and ${\cal E} \subseteq [p] \times [p]$,  the associated complex differential graph is ${\cal G}_\Delta = \left( V, {\cal E}_\Delta \right)$ where $\{i,j\} \in {\cal E}_\Delta$ iff $|[{\bm \Delta}]_{ij}| > 0$. Even though ${\bm \Delta}$ is Hermitian, $\hat{\bm \Delta}$ is not necessarily so. We make it Hermitian by setting $\hat{\bm \Delta}_{H} = \frac{1}{2} ( \hat{\bm \Delta} + \hat{\bm \Delta}^H)$, after obtaining $\hat{\bm \Delta}$.

\subsection{Optimization} \label{OPCGV}
Similar to \cite{Jiang2018} (also \cite{Yuan2017, Tugnait2024}), we use an alternating direction method of multipliers (ADMM) approach \cite{Boyd2010} with variable splitting to compute $\hat{\bm \Delta} = \arg\min_{\bm \Delta} L_\lambda({\bm \Delta}, \hat{\bm \Sigma}_x , \hat{\bm \Sigma}_y)$. Using variable splitting and adding the equality constraint ${\bm W}={\bm \Delta}$, consider
\begin{align} 
 \min_{\bm{\Delta} , {\bm W} } & \Big\{  L({\bm \Delta}, \hat{\bm \Sigma}_x , \hat{\bm \Sigma}_y) 
   + \lambda \sum_{k,\ell=1}^p |W_{k \ell}|  \Big\} \, , \label{eqn100} \\
				&	 \mbox{ subject to  }  \bm{\Delta} = {\bm W}  \, ,    \label{eqn100a}   
\end{align}
where in the penalty we use  ${\bm W}$ instead of ${\bm \Delta}$. Let ${\bm U} \in \mathbb{C}^{p \times p}$ denote the dual variable and $\rho >0$ denote the penalty parameter in the ADMM algorithm. The scaled augmented Lagrangian for this problem is \cite{Boyd2010}
\begin{align} 
 L_\rho&({\bm \Delta}, {\bm W}, {\bm U}) :=  L({\bm \Delta}, \hat{\bm \Sigma}_x , \hat{\bm \Sigma}_y) \\
  & + \lambda \sum_{k,\ell=1}^p |W_{k \ell}|   
					 + \frac{\rho}{2}  \| {\bm \Delta} - {\bm W} + {\bm U}\|^2_F  \, .  \label{eqn105}  
\end{align}

Given the $i$th iteration results $ \bm{\Delta}^{(i)}, {\bm W}^{(i)}, {\bm U}^{(i)}$, in the $(i+1)$st iteration, the ADMM algorithm executes the following 3 updates \cite{Boyd2010}:
\begin{itemize}
\item[(a)] $\bm{\Delta}^{(i+1)} \leftarrow \arg \min_{\bm{\Delta}} \, \bar{L}_a(\bm{\Delta})$ where
\[
  \bar{L}_a(\bm{\Delta}) := L({\bm \Delta}, \hat{\bm \Sigma}_x , \hat{\bm \Sigma}_y) + \frac{\rho}{2}  \| \bm{\Delta} 
						- {\bm W}^{(i)} + {\bm U}^{(i)}\|^2_F \, .
\]
\item[(b)] ${\bm W}^{(i+1)}  \leftarrow \arg \min_{ {\bm W} } \bar{L}_b({\bm W})$ where
\[
   \bar{L}_b({\bm W}) :=  \lambda \sum_{k, \ell=1}^p |W_{k \ell}| 
					+ \frac{\rho}{2}  \| \bm{\Delta}^{(i+1)} - {\bm W} + {\bm U}^{(i)} \|^2_F\, .
\]
\item[(c)] ${\bm U}^{(i+1)} \leftarrow {\bm U}^{(i)}  +
   \left( \bm{\Delta}^{(i+1)} - {\bm W}^{(i+1)} \right)$
\end{itemize}
Observe that $\bar{L}_a(\bm{\Delta})$ and $\bar{L}_b({\bm W})$ are convex in ${\bm \Delta}$ and ${\bm W}$, respectively.

We now address updates (a) and (b). \\
{\it Update (a)}. To minimize $\bar{L}_a(\bm{\Delta}) $ w.r.t.\ ${\bm \Delta}$, using Wirtinger calculus, we set 
\begin{align}
 {\bm 0}  & = \frac{\partial \bar{L}_a(\bm{\Delta})}{\partial {\bm \Delta}^\ast} \nonumber \\
   &=   
     \hat{\bm \Sigma}_x {\bm \Delta} \hat{\bm \Sigma}_y - (\hat{\bm \Sigma}_x-\hat{\bm \Sigma}_y) +
			\frac{\rho}{2} ( {\bm \Delta} - {\bm W}^{(i)} + {\bm U}^{(i)} ).  \label{eqn112}
\end{align}
Using $\mbox{vec}({\bm A} {\bm Y} {\bm B}) = ({\bm B}^\top \otimes {\bm A}) \mbox{vec}({\bm Y})$, we vectorize (\ref{eqn112}) to obtain
\begin{align} 
  &  \big(\hat{\bm \Sigma}_y^\ast \otimes \hat{\bm \Sigma}_x 
	        + \frac{\rho}{2} {\bm I}_p \otimes {\bm I}_p \big)  {\rm vec}({\bm \Delta}) \nonumber \\
	  & \quad    
		= {\rm vec} \big(\hat{\bm \Sigma}_x-\hat{\bm \Sigma}_y + \frac{\rho}{2} 
		  ({\bm W}^{(i)} - {\bm U}^{(i)}) \big) \, . \label{eqn116}
\end{align}
Direct matrix inversion solution of (\ref{eqn116}) requires inversion of a $p^2 \times p^2$ matrix. A computationally cheaper solution follows similar to that in \cite{Yuan2017, Jiang2018} where the real-valued case is addressed and here we consider complex-valued Hermitian matrices. Carry out eigen-decomposition of $\hat{\bm \Sigma}_x$ and $\hat{\bm \Sigma}_y$ as 
\begin{align}
  \hat{\bm \Sigma}_x & = {\bm Q}_x {\bm D}_x {\bm Q}_x^H \, , \quad {\bm Q}_x {\bm Q}_x^H = {\bm I}_p \, , 
	    \label{eqn117} \\
  \hat{\bm \Sigma}_y & = {\bm Q}_y {\bm D}_y {\bm Q}_y^H\, , \quad  {\bm Q}_y {\bm Q}_y^H = {\bm I}_p \, , \label{eqn118}
\end{align} 
where ${\bm D}_x$ and ${\bm D}_y$ are diagonal matrices. Define a matrix ${\bm D} \in \mathbb{R}^{p \times p}$ that organizes the diagonal of $({\bm D}_y \otimes {\bm D}_x + \frac{\rho}{2} {\bm I}_{p^2})^{-1}$ in a matrix with $(j,k)$th element as 
\begin{equation}
          [{\bm D}]_{jk} = \frac{1}{[{\bm D}_x]_{jj} [{\bm D}_y]_{kk} + \frac{\rho}{2}} \; ,  \label{eqn121}
\end{equation} 
and recall that the symbol $\circ$ denotes the Hadamard matrix product. 
 Then $\hat{\bm \Delta}$ that minimizes $\hat{L}_a(\bm{\Delta})$ is given by (the derivation of (\ref{eqn120}) is given in Appendix \ref{append1})
\begin{align} 
  \hat{\bm \Delta} = & {\bm Q}_x \Big[ {\bm D} \circ [{\bm Q}_x^H \big(\hat{\bm \Sigma}_x-\hat{\bm \Sigma}_y 
	         + \frac{\rho}{2} ( {\bm W}^{(i)} - {\bm U}^{(i)}) \big)
		   {\bm Q}_y ] \Big] {\bm Q}_y^H  \label{eqn120}  
\end{align}
Note that the eigen-decomposition of $\hat{\bm \Sigma}_x$ and $\hat{\bm \Sigma}_y$ has to be done only once. With ${\bm A}^{(i)} = {\bm W}^{(i)} - {\bm U}^{(i)}$, we have
\begin{align} 
  {\bm \Delta}^{(i+1)} = & {\bm Q}_x \Big[ {\bm D} \circ [{\bm Q}_x^H \big(\hat{\bm \Sigma}_x-\hat{\bm \Sigma}_y 
	     + \frac{\rho}{2} {\bm A}^{(i)} \big) {\bm Q}_y ] \Big] {\bm Q}_y^H  \, . \label{eqn125}  
\end{align}

{\it Update (b)}. Notice that $\bar{L}_b({\bm W})$ is separable in $(k, \ell)$ with 
\begin{align} 
  \bar{L}_b(W_{k \ell}) & =  \lambda |W_{k \ell}| +\frac{\rho}{2}  
	  |{\Delta}_{k \ell}^{(i+1)} - { W}_{k \ell} + { U}_{k \ell}^{(i)}|^2 \, , \\
  \bar{L}_b({\bm W}) & =  \sum_{k, \ell=1}^p \bar{L}_b(W_{k \ell}) \, .
\end{align} 
Following \cite[Lemma 1]{Tugnait2022} and using $(a)_+ = \max(0,a)$, $\bar{L}_b(W_{k \ell})$ is minimized  by the lasso solution 
\begin{align} 
  & W_{k \ell}^{(i+1)}  = \Big( 1 - \frac{(\lambda / \rho)}{  | [{\bm \Delta}^{(i+1)}+ {\bm U}^{(i)}]_{k \ell} | } \Big)_+    
								  [{\bm \Delta}^{(i+1)}+ {\bm U}^{(i)}]_{k \ell} \, . \label{eqn130}  
\end{align}

\subsubsection{Convergence} \label{conv1}
A stopping (convergence) criterion following \cite[Sec.\ 3.3.1]{Boyd2010} can be devised. The stopping criterion is based on primal and dual residuals being small where, in our case, at $(i+1)$st iteration, the primal residual is given by $\bm{\Delta}^{(i+1)} - {\bm W}^{(i+1)}$ and the dual residual by $\rho ({\bm W}^{(i+1)} - {\bm W}^{(i)})$. The convergence criterion is met when the norms of these residuals are below some threshold. 

The objective function $L_\lambda({\bm \Delta}, \hat{\bm \Sigma}_x , \hat{\bm \Sigma}_y)$, given by (\ref{eqn25}), is convex. It is also closed, proper and lower semi-continuous. Hence, for any fixed $\rho > 0$, the ADMM algorithm is guaranteed to converge \cite[Sec.\ 3.2 and Appendix A]{Boyd2010}, in the sense that we have primal residual convergence to 0, dual residual convergence to 0, and the objective function convergence to the optimal value.

\section{Differential Time Series Graphs: System Model} \label{SM} 
We now turn to the general problem of differential times series graph estimation. 
Consider two independent stationary (real-valued), zero-mean,  $p-$dimensional multivariate Gaussian time series ${\bm x}(t)$ and ${\bm y}(t)$, $t \in \mathbb{Z}$, with PSDs ${\bm S}_x(f) \succ {\bm 0}$ and ${\bm S}_y(f) \succ {\bm 0}$, respectively, for every $f \in [0,1]$, and the CIGs ${\cal G}_x = \left( V, {\cal E}_x \right)$ and ${\cal G}_y = \left( V, {\cal E}_y \right)$, respectively. As discussed earlier in Sec.\ \ref{intro}, the edge $\{ i,j \} \not\in {\cal E}_x$ iff  $[{\bm S}_x^{-1}(f)]_{ij} \equiv 0$ and $\{ i,j \} \not\in {\cal E}_y$ iff  $[{\bm S}_y^{-1}(f)]_{ij} \equiv 0$ \cite{Dahlhaus2000}. In differential network analysis for time-dependent data, one is interested in the difference ${\bm \Delta}(f) = {\bm S}_y^{-1}(f) - {\bm S}_x^{-1}(f)$, and in the associated differential graph ${\cal G}_\Delta = \left( V, {\cal E}_\Delta \right)$ we have  $\{i,j\} \not\in {\cal E}_\Delta$ iff $[{\bm \Delta}(f)]_{ij} = 0$ for every $f \in [0,1]$.

Given data $\{ {\bm x}(t) \}$ and $\{ {\bm y}(t) \}$, our objective is to first estimate the inverse PSDs ${\bm S}_x^{-1}(f)$ and ${\bm S}_y^{-1}(f)$ at distinct frequencies, and then select the edge $\{i,j\}$ in the differential time series graph ${\cal G}_\Delta$ based on whether or not $[{\bm \Delta}(f)]_{ij} = 0$ for every $f \in [0,1]$.

\subsection{Problem Formulation} \label{BPF1}
Given time-domain data $\{ {\bm x}(t) \}_{t=1}^{n_x}$ and $\{ {\bm y}(t) \}_{t=1}^{n_y}$ originating from two independent stationary, zero-mean,  multivariate Gaussian time series ${\bm x}(t) \in \mathbb{R}^p$ and ${\bm y}(t) \in \mathbb{R}^p$. For simplicity, we take $n_x=n_y=n$. With $\iota = \sqrt{-1}$, define the (normalized) discrete Fourier transforms (DFTs) 
\begin{align} 
   {\bm d}_x(f_m) = & \frac{1}{\sqrt{n}} \sum_{t=1}^{n} {\bm x}(t) \exp \left( - \iota 2 \pi f_m (t-1) \right) \, , 
	        \label{app1eq50} 
\end{align}
\begin{align} 
	  {\bm d}_y(f_m) = & \frac{1}{\sqrt{n}} \sum_{t=1}^{n} {\bm y}(t) \exp \left( - \iota 2 \pi f_m (t-1) \right) \, , 
	        \label{app1eq50a} \\
					f_m = & \frac{m}{n}, \; m=0,1, \cdots , n-1. \label{app1eq50b} 
\end{align} 
The set of complex random vectors $\{{\bm d}_x(f_m), \, {\bm d}_y(f_m)\}_{m=0}^{n/2}$ is a sufficient statistic for any inference problem based on data set $\{ {\bm x}(t), \, {\bm y}(t) \}_{t=1}^{n}$ since given $\{{\bm d}_x(f_m), \, {\bm d}_y(f_m)\}_{m=0}^{n/2}$, one can recover $\{ {\bm x}(t), \, {\bm y}(t) \}_{t=1}^{n}$ via inverse DFT. \cite{Tugnait2022}.

{\subsubsection{Model Assumptions}} \label{MAss}
We assume the following:
\begin{itemize}
\setlength{\itemindent}{0.1in}
\item[(A1)] The time series $\{ {\bm x}(t) \}_{t \in \mathbb{Z}}$ is zero-mean stationary and Gaussian, satisfying 
\[
    \sum_{\tau = -\infty}^\infty | [{\bm R}_{xx}( \tau )]_{k \ell} | < \infty \mbox{ for every } 
		  k, \ell \in V = [p]  \, ,
\] 
and similarly for $\{ {\bm y}(t) \}_{t \in \mathbb{Z}}$.
\item[(A2)] For some integer $m_t > 0$, let $K=2m_t+1$. Pick 
\begin{align*}
 M = & \left\lfloor (\frac{n}{2}-m_t-1)/K \right\rfloor \, , \\
 \tilde{f}_k = & ((k-1)K+m_t+1)/n \; \mbox{ for } \; k \in [M] \, ,
\end{align*}
yielding $M$ equally spaced frequencies $\tilde{f}_k$ in the interval $(0,0.5)$. Assume that for $\ell = -m_t, -m_t+1, \cdots , m_t$, the PSD matrices ${\bm S}_x(f)$ and ${\bm S}_y(f)$ satisfy 
\begin{align}  
   {\bm S}_x & (\tilde{f}_{k,\ell}) =  {\bm S}_x(\tilde{f}_k) \,  , 
	\;\; {\bm S}_y  (\tilde{f}_{k,\ell}) =  {\bm S}_y(\tilde{f}_k) \,  ,\label{eqth1_160}  \\
	 \mbox{ where } \;  &
	  \tilde{f}_{k,\ell} = \big((k-1)K+m_t+1 + \ell \big)/n \, .
\end{align}
\end{itemize}
Assumption (A1) is needed to invoke \cite[Theorem 4.4.1]{Brillinger} regarding distribution of the DFTs $\{{\bm d}_x(f_m), \, {\bm d}_y(f_m)\}_{m=0}^{n/2}$. Assumption (A2) is a local smoothness assumption which implies that ${\bm S}_x(f_k)$ and ${\bm S}_y(f_k)$ are constant over $K=2m_t+1$ consecutive frequency points $f_m$'s, $m_t > 0$.  

It turns out that for ``large'' $n$, under assumption (A1), the DFT ${\bm d}_x(f_m)$ is a complex-valued proper (i.e., circularly symmetric) Gaussian vector $\sim {\mathcal N}_c( {\bf 0}, {\bm S}_x(f_m))$, and (mutually) independent for $m = 1,2, \cdots, (n/2)-1$, ($n$ even) \cite[Theorem 4.4.1]{Brillinger}, though not identically distributed. Also, ${\bm d}_x(f_0)$ and ${\bm d}_x(f_{n/2})$ ($n$ even) are real-valued Gaussian vectors. Similar comments apply to ${\bm d}_y(f_m)$. We exclude the frequency points $f_0$ and $f_{n/2}$ from further consideration. Define
\begin{align}
\hat{\bm S}_{xk} & = \frac{1}{K} \sum_{\ell= - m_t}^{m_t} {\bm d}_x(\tilde{f}_{k,\ell}) 
		 {\bm d}_x^H(\tilde{f}_{k,\ell}) \, , \label{specx} \\
\hat{\bm S}_{yk} & = \frac{1}{K} \sum_{\ell= - m_t}^{m_t} {\bm d}_y(\tilde{f}_{k,\ell}) 
		 {\bm d}_y^H(\tilde{f}_{k,\ell})  \label{specy}
\end{align}
where $\hat{\bm S}_{xk}$ and $\hat{\bm S}_{yk}$ represent PSD estimators at frequency $\tilde{f}_{k}$ using unweighted frequency-domain smoothing \cite{Brillinger}.  By the local smoothness assumption (A2), for $\ell = -m_t, -m_t+1, \cdots , m_t$, we have
\begin{align}
 {\bm d}_x(\tilde{f}_{k,\ell}) & \sim {\mathcal N}_c( {\bm 0}, {\bm S}_x(\tilde{f}_{k}) ) \, , \\
 {\bm d}_y(\tilde{f}_{k,\ell}) & \sim {\mathcal N}_c( {\bm 0}, {\bm S}_y(\tilde{f}_{k}) ) \, .
\end{align}

{\subsubsection{Multiple Complex Differential Graphs}} \label{MCGM}
To lighten notation, henceforth we will denote the true values of ${\bm S}_x(\tilde{f}_{k})$ and ${\bm S}_y(\tilde{f}_{k})$ as ${\bm S}_{xk}^\diamond$ and ${\bm S}_{yk}^\diamond$, respectively, with their respective sample estimates $\hat{\bm S}_{xk}$ and $\hat{\bm S}_{yk}$, $k \in [M]$. 

Our objective is to ascertain if ${\bm \Delta}(f) = {\bm S}_y^{-1}(f) - {\bm S}_x^{-1}(f) = {\bm 0}$ $\forall f \in [0,1]$. Since the PSD matrix ${\bm S}(f)$ of any real discrete-time zero-mean stationary random process is periodic with period one and ${\bm S}(-f) = {\bm S}^H(f)$, it is enough to check if ${\bm \Delta}(f) = {\bm 0}$ $\forall f \in [0,0.5]$, and therefore, in the associated differential graph ${\cal G}_\Delta = \left( V, {\cal E}_\Delta \right)$ we have  $\{i,j\} \not\in {\cal E}_\Delta$ iff $[{\bm \Delta}(f)]_{ij} = 0$ for every $f \in [0,0.5]$. Let ${\bm \Delta}_{k} = {\bm S}_{yk}^{-1} - {\bm S}_{xk}^{-1}$ and ${\bm \Delta}_{k}^\diamond = ({\bm S}_{yk}^\diamond)^{-1} - ({\bm S}_{xk}^\diamond)^{-1}$. Under assumption (A2) (and recalling that we exclude frequency points $f_0$ and $f_{n/2}$), ${\bm \Delta}(f), ~\forall f \in [0,0.5]$ translates to ${\bm \Delta}_{k}, ~ k \in [M]$ such that $\{ {\bm \Delta}(f) = {\bm 0}, ~\forall f \in [0,0.5] \}$ is equivalent to $\{ {\bm \Delta}_{k}={\bm 0}, ~ k \in [M] \}$, and therefore, in the associated differential graph ${\cal G}_\Delta = \left( V, {\cal E}_\Delta \right)$ we have  $\{i,j\} \not\in {\cal E}_\Delta$ iff $[{\bm \Delta}_{k}]_{ij} = 0$ for every $k \in [M]$. 

Observe that for any fixed $k \in [M]$, ${\bm \Delta}_{k}^\diamond$ characterizes a complex differential graph (cf.\ Sec.\ \ref{CDG}) with ``precision matrix'' difference $({\bm S}_{yk}^\diamond)^{-1} - ({\bm S}_{xk}^\diamond)^{-1}$.  To estimate ${\bm \Delta}_{k}^\diamond$ we have $K=2m_t+1$ independent complex measurements ${\bm d}_x(\tilde{f}_{k,\ell})$ and ${\bm d}_y(\tilde{f}_{k,\ell})$, $\ell = -m_t, -m_t+1, \cdots , m_t$. The D-trace loss for the $k$th differential graph is $L({\bm \Delta}_k, \hat{\bm S}_{xk} , \hat{\bm S}_{yk})$ with $L(\cdot,\cdot,\cdot)$ specified by (\ref{deqn200}). Moreover, $L({\bm \Delta}_{k}, {\bm S}_{xk}^\diamond , {\bm S}_{yk}^\diamond)$ is strictly convex in ${\bm \Delta}_{k}$ with a unique minimum at ${\bm \Delta}_{k}^\diamond$ (see Sec.\ \ref{PCGV}).

\subsubsection{D-Trace Loss} \label{DTL} 
Now we wish to ascertain if $[{\bm \Delta}_{k}]_{ij} = 0$ for every $k \in [M]$ which calls for joint consideration of all $M$ complex differential graphs. Define
\begin{align}  
	\tilde{\bm \Delta} := & [ {\bm \Delta}_1 , \; {\bm \Delta}_2 , \; \cdots , \; {\bm \Delta}_M] \in \mathbb{C}^{p \times p M} \, . \label{app1eq10001} 
\end{align}
In order to exploit every $k \in [M]$, we propose the cost 
\begin{align} 
  \tilde{L}(\tilde{\bm \Delta}) = & \sum_{k=1}^M L({\bm \Delta}_k, \hat{\bm S}_{xk} , \hat{\bm S}_{yk}) \, . \label{app1eq1000} 
\end{align}
Define
\begin{align}
	\tilde{\bm \Delta}^{(ij)} := & \big[[{\bm \Delta}_1]_{ij} , \, [{\bm \Delta}_2]_{ij} , \, 
	  \cdots, \, [{\bm \Delta}_M]_{ij} \big]^\top \in \mathbb{C}^M \, . \label{app1eq1002}
\end{align}
Substitute $\hat{\bm S}_{yk}={\bm S}_{yk}^\diamond$ and $\hat{\bm S}_{xk}={\bm S}_{xk}^\diamond$ in $\tilde{L}(\tilde{\bm \Delta})$ and denote it by $\tilde{L}(\tilde{\bm \Delta}^\diamond)$. Then $\tilde{L}(\tilde{\bm \Delta}^\diamond)$, being a sum of strictly convex functions (cf.\ Sec.\ \ref{PCGV}), is strictly convex and has a unique minimum at ${\bm \Delta}_{k} = {\bm \Delta}_{k}^\diamond$, $k \in [M]$ (cf.\ Sec.\ \ref{PCGV}). Since data-based $\tilde{L}(\tilde{\bm \Delta})$ is a sum of convex functions $L({\bm \Delta}_k, \hat{\bm S}_{xk} , \hat{\bm S}_{yk})$ (cf.\ Sec.\ \ref{PCGV}), it is convex, but not necessarily strictly convex. 

Since true values ${\bm \Delta}_{k}^\diamond$'s are unavailable, our objective then is to estimate $\tilde{\bm \Delta}$ by minimizing data-based $\tilde{L}(\tilde{\bm \Delta})$ with resulting estimate 
\begin{align}
  \hat{\tilde{\bm \Delta}} & =  [ \hat{\bm \Delta}_1 , \; \hat{\bm \Delta}_2 , \; \cdots , \; \hat{\bm \Delta}_M] \, .
\end{align}
We estimate the edgeset ${\cal E}_\Delta$ of the differential time-series graph as 
\begin{align}
  \hat{\cal E}_\Delta = \Big\{ \{i,j\} \, : \, \|\hat{\tilde{\bm \Delta}}^{(ij)} \| > 0 \Big\} \, .
\end{align}

\section{Penalized D-Trace Loss} \label{PDTL}
In the high-dimensional case of $K < p$, to enforce sparsity and to make the problem well-conditioned, we propose to minimize a group penalized version of $\tilde{L}(\tilde{\bm \Delta})$ w.r.t.\ ${\bm \Delta}_k$s, given by 
\begin{equation}
  L_f(\tilde{\bm \Delta}) = \tilde{L}(\tilde{\bm \Delta})
	   + \sum_{i, j=1}^p h_\lambda \left( \| \tilde{\bm \Delta}^{(ij)} \| \right)
	  \label{deqn210}
\end{equation}
where, for $u \in \mathbb{R}$, $h_\lambda(u)$ is a penalty function that is a function of $|u|$. The penalty operates on the group $\tilde{\bm \Delta}^{(ij)} \in \mathbb{C}^M$, instead of individual elements of $\tilde{\bm \Delta}$. 
 
The following penalty functions are considered: 
\begin{itemize}
\item[(1)] {\it Lasso}. For some $\lambda > 0$,  $h_\lambda(u) = \lambda |u|$, $u \in \mathbb{R}$. It is a convex function that is widely used. 
\item[(2)] {\it Log-sum}. For some $\lambda > 0$ and $1 \gg \epsilon > 0$, $h_\lambda(u) = \lambda \epsilon \, \ln \left( 1 + \frac{|u|}{\epsilon} \right)$. It is a nonconvex function. 
\item[(3)] {\it SCAD}. For some $\lambda > 0$ and $a > 2$, $h_\lambda(u) = \lambda | u |$ for $|u| \le \lambda$, $= (2 a \lambda | u |- | u |^2 - \lambda^2)/(2 (a-1))$ for $\lambda < |u| < a \lambda$, and $=\lambda^2 (a+1)/2$ for $|u| \ge a$. It is a nonconvex function.
\end{itemize}
In \cite{Loh2017} the log-sum penalty is defined as $h_\lambda(u) = \ln (1+\lambda |u|)$ whereas in \cite{Candes2008}, it is defined as $h_\lambda(u) = \lambda \, \ln \left( 1 + \frac{|u|}{\epsilon} \right)$. We follow  \cite{Candes2008} but modify it so that, as for the lasso and SCAD penalties, our $h_\lambda(u)$ yields $\lim_{u \rightarrow 0^+} h_\lambda^\prime(u) = \lambda$ where $h_\lambda^\prime(u) := \frac{d h_\lambda (u)}{du}$.

\section{Optimization} \label{OPT}
The objective function $L_f(\tilde{\bm \Delta})$ is  non-convex in ${\bm \Delta}$ for the non-convex SCAD and log-sum penalties, and convex for the lasso penalty. Now we discuss an ADMM approach, following the ADMM approach discussed in Sec.\ \ref{OPCGV} for lasso, to attain a local minimum of $L_f(\tilde{\bm \Delta})$ for the non-convex SCAD and log-sum penalties, and a global minimum for the lasso penalty.

For non-convex $h_\lambda(u)$, we use a local linear approximation (LLA) (as in \cite{Zou2008, Lam2009}), to yield 
\begin{equation}
  h_{\lambda}(u) \approx h_{\lambda}(|u_0|) 
	 + h_\lambda^\prime(|u_0|) (|u| - |u_0|)
	 \, \rightarrow \, h_\lambda^\prime(|u_0|) |u| \, ,
\end{equation}
where $h^\prime (x) = dh(x)/dx$, $u_0$ is an initial guess, and the gradient of the penalty function is 
\begin{align} 
  h_\lambda^\prime(|u_0|) = & \left\{ \begin{array}{l}
							\frac{\lambda \epsilon }{|u_0| +\epsilon}   \mbox{ for log-sum}, \\ 
			 \left\{ \begin{array}{ll} \lambda , & \mbox{if  } |u_0| \le \lambda \\
				    \frac{ a \lambda - | u_0 |}{ a-1} , 
									& \mbox{if  } \lambda < |u_0| \le a \lambda \\ 
						0 , & \mbox{if  } a \lambda < |u_0|  \end{array} \right. \\
						  \quad\quad \mbox{ for SCAD}. \end{array} \right.  
\end{align}
Therefore, with $u_0$ fixed, we need to consider only the term dependent upon $u$ for optimization w.r.t.\ $u \,$:
\begin{equation}
  h_{\lambda}(u)  \, \rightarrow \, h_\lambda^\prime(|u_0|) \, |u| \, .
\end{equation}
By \cite[Theorem 1]{Zou2008}, the LLA provides a majorization of the non-convex penalty, thereby yielding a majorization-minimization approach. By \cite[Theorem 2]{Zou2008}, the LLA is the best convex majorization of the LSP and SCAD penalties. 

With some initial guess $\bar{\tilde{\bm \Delta}} =  [ \bar{\bm \Delta}_1 , \; \bar{\bm \Delta}_2 , \; \cdots , \; \bar{\bm \Delta}_M]$, in LSP we replace 
\begin{align}
  h_\lambda \left( \| \tilde{\bm \Delta}^{(ij)} \| \right)  & \rightarrow \lambda_{ij} :=
	 \frac{\lambda \epsilon }{\| \bar{\tilde{\bm \Delta}}^{(ij)} \|  +\epsilon} \, .  \label{logsumInit}
\end{align}
The solution $\hat{\tilde{\bm \Delta}}_{\rm lasso}$ to the convex lasso-penalized objective function may be used as an initial guess with $\bar{\tilde{\bm \Delta}} = \hat{\tilde{\bm \Delta}}_{\rm lasso}$. Similarly, for SCAD, we have 
\begin{align} \label{scadInit}
  \lambda_{ij} = &
			 \left\{ \begin{array}{ll} \lambda , & \mbox{if  } \| \bar{\tilde{\bm \Delta}}^{(ij)} \| \le \lambda \\
				    \frac{ a \lambda - \| \bar{\tilde{\bm \Delta}}^{(ij)} \|}{ a-1} , 
									& \mbox{if  } \lambda < \| \bar{\tilde{\bm \Delta}}^{(ij)} \| \le a \lambda \\ 
						0 , & \mbox{if  } a \lambda < \| \bar{\tilde{\bm \Delta}}^{(ij)} \|  \end{array} \right.  \, .  
\end{align}
With LLA, the original objective function is transformed to its convex LLA approximation
\begin{align} 
  \tilde{L}_f(\tilde{\bm \Delta}) & = \tilde{L}(\tilde{\bm \Delta})
	   + \sum_{i, j=1}^p \lambda_{ij}  \| \tilde{\bm \Delta}^{(ij)} \|  \, .
	  \label{admm}
\end{align}
For lasso, we have $\lambda_{ij} = \lambda$ $\forall i,j$. If $\bar{\tilde{\bm \Delta}}^{(ij)} = {\bm 0}$, we obtain $\lambda_{ij} = \lambda$ $\forall i,j$.

We follow an ADMM approach following the ADMM approach discussed in Sec.\ \ref{OPCGV} for lasso, for both lasso and LLA to LSP/SCAD. For non-convex penalties, we have an iterative solution: first solve with lasso penalty, then use this solution for initialization to LLA and solve again the LLA convex problem. In practice, just two iterations seem to be enough.
Using variable splitting and adding the equality constraint $\tilde{\bm W}=\tilde{\bm \Delta}$ with $\tilde{\bm W}= [ {\bm W}_1 \; \cdots \; {\bm W}_M]$, ${\bm W}_k \in \mathbb{C}^{p \times p}$ for $k \in [M]$, consider ($\tilde{\bm W}^{(ij)}$ is defined similar to (\ref{app1eq1002}))
\begin{align} 
 \min_{\tilde{\bm \Delta} , \tilde{\bm W} } & \Big\{ \tilde{L}_f(\tilde{\bm \Delta}) 
   + \sum_{i, j=1}^p \lambda_{ij}  \| \tilde{\bm W}^{(ij)} \|  \Big\} \, , \label{1eqn100} \\
				&	 \mbox{ subject to  }  \tilde{\bm \Delta} = \tilde{\bm W}  \, ,    \label{1eqn100a}   
\end{align}
where, in the penalty, we use  $\tilde{\bm W}$ instead of $\tilde{\bm \Delta}$. Let $\tilde{\bm U}= [ {\bm U}_1 \; \cdots \; {\bm U}_M]$, ${\bm U}_k \in \mathbb{C}^{p \times p}$ for $k \in [M]$, denote the dual variables and $\rho >0$ denote the penalty parameter in the ADMM algorithm. The scaled augmented Lagrangian for this problem is \cite{Boyd2010}
\begin{align} 
&  \tilde{L}_\rho(\tilde{\bm \Delta}, \tilde{\bm W}, \tilde{\bm U}) =  \tilde{L}_f(\tilde{\bm \Delta})  
   +  \sum_{i, j=1}^p \lambda_{ij}  \| \tilde{\bm W}^{(ij)} \| \nonumber \\
  & \quad \quad \quad   
					 + \frac{\rho}{2} \sum_{k=1}^M \| {\bm \Delta}_k - {\bm W}_k + {\bm U}_k \|^2_F \, .  \label{feqn105}  
\end{align}

Following Sec.\ \ref{OPCGV}, given the results $ \tilde{\bm \Delta}^{(m)}, \tilde{\bm W}^{(m)}, \tilde{\bm U}^{(m)}$ of the $m$th iteration, in the $(m+1)$st iteration, an ADMM algorithm executes the following three updates:
\begin{itemize}
\item[(i)] $\tilde{\bm \Delta}^{(m+1)} \leftarrow \arg \min _{\tilde{\bm \Delta}} \sum_{k=1}^M \tilde{L}_{ak} ( {\bm \Delta}_k )$ where
\[ 
\tilde{L}_{ak} ( {\bm \Delta}_k) = L({\bm \Delta}_k, \hat{\bm S}_{xk} , \hat{\bm S}_{yk}) + 
  \frac{\rho}{2}  \| {\bm \Delta}_k - {\bm W}_k^{(m)} + {\bm U}_k^{(m)} \|^2_F .
\]
\item[(ii)] $\tilde{\bm W}^{(m+1)} \leftarrow \arg \min _{\tilde{\bm W}} \tilde{L}_b(\tilde{\bm W})$ where
\begin{align*}       
     \tilde{L}_b(\tilde{\bm W}) = & 
				 \frac{\rho}{2} \sum_{k=1}^M \| {\bm \Delta}_k^{(m+1)} - {\bm W}_k + {\bm U}_k^{(m)} \|^2_F  \\
				& \quad + \sum_{i, j=1}^p \lambda_{ij} \| \tilde{\bm W}^{(ij)} \| \, .
\end{align*}
\item[(iii)] $\tilde{\bm U}^{(m+1)}  \leftarrow \tilde{\bm U}^{(m)}  +
    \tilde{\bm \Delta}^{(m+1)} - \tilde{\bm W}^{(m+1)} $.
\end{itemize}

Some details regarding updates (i) and (ii) are given below. \\
\noindent {\it Update (i)}: The optimization in step (i) is separable in ${\bm \Delta}_k$, and the solution discussed in Sec.\ \ref{OPCGV} applies. Perform the eigen-decomposition of $\hat{\bm S}_{xk}$ and $\hat{\bm S}_{yk}$ as $\hat{\bm S}_{xk} = {\bm Q}_{xk} {\bm D}_{xk} {\bm Q}_{xk}^H$ and $\hat{\bm S}_{yk} = {\bm Q}_{yk} {\bm D}_{yk} {\bm Q}_{yk}^H$ where ${\bm D}_{xk}$ and ${\bm D}_{yk}$ are diagonal matrices,  ${\bm Q}_{xk} {\bm Q}_{xk}^H = {\bm I}_p$ and ${\bm Q}_{yk} {\bm Q}_{yk}^H = {\bm I}_p$. Define a matrix ${\bm D}^{(k)} \in \mathbb{R}^{p \times p}$ that organizes the diagonal of $({\bm D}_{yk} \otimes {\bm D}_{xk} + \frac{\rho}{2} {\bm I}_{p^2})^{-1}$ in a matrix with the $(i,j)$th element $[{\bm D}^{(k)}]_{ij} = 1/([{\bm D}_{xk}]_{ii} [{\bm D}_{yk}]_{jj} + \frac{\rho}{2})$. Then, for $k \in [M]$, we have 
\begin{align} 
  {\bm \Delta}_k^{(m+1)} = & {\bm Q}_{xk} \Big[ {\bm D}^{(k)} 
	   \circ \big[{\bm Q}_{xk}^H \big(\hat{\bm S}_{xk}-\hat{\bm S}_{yk}  \nonumber \\
	   &  + \frac{\rho}{2} ({\bm W}_k^{(m)} - {\bm U}_k^{(m)}) \big) {\bm Q}_{yk} \big] \Big] 
		    {\bm Q}_{yk}^H \, . \label{feqn125} 
\end{align}

\noindent {\it Update (ii)}: The optimization in step (ii) is separable in $\tilde{\bm W}^{(ij)}$, and the solution discussed in Sec.\ \ref{OPCGV} applies with $\lambda \rightarrow \lambda_{ij}$. With ${\bm A}_k = {\bm \Delta}_k^{(m+1)} + {\bm U}_k^{(m)}$, $(b)_+ = \max(0,b)$, and for $k\in[M]$ and $i,j \in [p]$,
\begin{align} 
	& [{\bm W}_k^{(m+1)}]_{ij} = \left( 1 - \frac{\lambda_{ij}}
			{ \rho \| \tilde{\bm A}^{(ij)}\|}
			   \right)_+ [{\bm A}_k]_{ij} \; ,   \label{feqn130} \\
 \mbox{where } &   \tilde{\bm A}^{(ij)} = \big[[{\bm A}_1]_{ij} , \, \cdots, \, [{\bm A}_M]_{ij} \big]^\top \in \mathbb{C}^M \, . \label{feqn131}  
\end{align}

\begin{algorithm} [htbp]
\caption{ADMM Algorithm for Solving (\ref{feqn105})}
\label{alg0}

\algorithmicrequire{\; PSD estimators $\hat{\bm S}_{xk}$ and $\hat{\bm S}_{yk}$, $k \in [M]$ (computed using (\ref{specx}) and (\ref{specy})), regularization and penalty parameters $\lambda_{ij}$ ($i,j \in [p]$) and $\rho=\bar{\rho}$, tolerances $\tau_{abs}$ and $\tau_{rel}$, variable penalty factor $\bar{\mu}$, maximum number of iterations $m_{\max}$}. Initial guess $\bar{\bm \Delta}_k$, $k \in [M]$. \\
\algorithmicensure{\;\ Estimated $\hat{\bm \Delta}_k$, $k \in [M]$.}

\begin{algorithmic}[1] 
\STATE Initialize: ${\bm \Delta}_k^{(0)} = \bar{\bm \Delta}_k$, ${\bm U}_k^{(0)} = {\bm W}_k^{(0)} = {\bm 0}$, $k \in [M]$, $\rho^{(0)} = \bar{\rho}$ 
\STATE Eigen-decompose  $\hat{\bm S}_{xk}$ and $\hat{\bm S}_{yk}$ as $\hat{\bm S}_{xk} = {\bm Q}_{xk} {\bm D}_{xk} {\bm Q}_{xk}^H$ and $\hat{\bm S}_{yk} = {\bm Q}_{yk} {\bm D}_{yk} {\bm Q}_{yk}^H$, $k \in [M]$.
\STATE converged = \FALSE, $m=0$
\WHILE{converged = \FALSE $\;$ \AND $\;$ $m \le m_{\max}$,}
\STATE For $k \in [M]$, construct ${\bm D}^{(k)} \in \mathbb{R}^{p \times p}$  with $[{\bm D}^{(k)}]_{ij} = 1/([{\bm D}_{xk}]_{ii} [{\bm D}_{yk}]_{jj} + \frac{\rho^{(m)}}{2})$.
\STATE For $k \in [M]$, set ${\bm \Delta}_k^{(m+1)} =  {\bm Q}_{xk} \Big[ {\bm D}^{(k)} 
	   \circ \big[{\bm Q}_{xk}^H \big(\hat{\bm S}_{xk}-\hat{\bm S}_{yk}  
	     + \frac{\rho}{2} ({\bm W}_k^{(m)} - {\bm U}_k^{(m)}) \big) {\bm Q}_{yk} \big] \Big] {\bm Q}_{yk}^H$.
\STATE For $k \in [M]$, define ${\bm A}_k = {\bm \Delta}_k^{(m+1)} + {\bm U}_k^{(m)}$ and $\tilde{\bm A}^{(ij)} = \big[[{\bm A}_1]_{ij} , \, \cdots, \, [{\bm A}_M]_{ij} \big]^\top$. For $k\in[M]$ and $i,j \in [p]$,
$[{\bm W}_k^{(m+1)}]_{ij} = \left( 1 - \frac{\lambda_{ij}}
			{ \rho^{(m)} \, \| \tilde{\bm A}^{(ij)}\|}
			   \right)_+ [{\bm A}_k]_{ij}$. 
\STATE Dual update ${\bm U}_k^{(m+1)}  \leftarrow {\bm U}_k^{(m)}  +
    {\bm \Delta}_k^{(m+1)} - {\bm W}_k^{(m+1)} $, $k \in [M]$. 
\STATE Check convergence. With $e_1$, $e_2$, $e_3$, ${\bm R}_p^{(m+1)}$, ${\bm R}_d^{(m+1)}$, $\tau_{pri}$ and $\tau_{dual}$ as defined in (\ref{alg100})-(\ref{alg160}), respectively, let $d_p =  \|{\bm R}_p^{(m+1)}\|_F$ and $d_d = \|{\bm R}_d^{(m+1)}\|_F$. 
If $( d_p \le \tau_{pri}) \; \AND \; (d_d \le \tau_{dual})$, set converged = \TRUE .
\STATE Update penalty parameter $\rho$ $\,$ : 
\[
  \rho^{(m+1)} = \left\{ \begin{array}{ll} 2 \rho^{(m)} & \mbox{if  } d_p > \bar{\mu} \, d_d \\  
	                                         \rho^{(m)} /2 & \mbox{if  } d_d > \bar{\mu} \, d_p \\
																					 \rho^{(m)} & \mbox{otherwise} \, . \end{array} \right.
\]
We also need to set ${\bm U}^{(m+1)} = {\bm U}^{(m+1)}/2$ for $d_p > \bar{\mu} d_d$ and ${\bm U}^{(m+1)} = 2 {\bm U}^{(m+1)}$ for $d_d > \bar{\mu} d_p$.
\STATE $m \leftarrow m+1$
\ENDWHILE
\STATE With ${\bm \Delta}_k$, $k \in [M]$, denoting the converged estimates, set $\hat{\bm \Delta}_k = ({\bm \Delta}_k+{\bm \Delta}_k^H)/2$, $k \in [M]$, and 
\[ 
   \hat{\tilde{\bm \Delta}} = [ \hat{\bm \Delta}_1 , \; \hat{\bm \Delta}_2 , \; \cdots , \; \hat{\bm \Delta}_M] \,  .
\]  
\end{algorithmic}
\end{algorithm}

A pseudocode for the ADMM algorithm to solve (\ref{feqn105}) is given in Algorithm \ref{alg0} where we use the stopping (convergence) criterion following \cite[Sec.\ 3.3.1]{Boyd2010} and varying penalty parameter $\rho$ following \cite[Sec.\ 3.4.1]{Boyd2010}. The variables defined in (\ref{alg100})-(\ref{alg160}) are needed in Algorithm \ref{alg0} with ${\bm \Delta}_k^{(m+1)}$, ${\bm W}_k^{(m+1)}$, ${\bm U}_k^{(m+1)}$ as defined therein:
\begin{align}
& e_1 =  \|[{\bm \Delta}_1^{(m+1)}, \cdots , {\bm \Delta}_M^{(m+1)}]\|_F  \label{alg100} \\
& e_2 = \|[{\bm W}_1^{(m+1)}, \cdots , {\bm W}_M^{(m+1)}]\|_F \label{alg110} \\
& e_3 =  \|[{\bm U}_1^{(m+1)}, \cdots , {\bm U}_M^{(m+1)}]\|_F  \label{alg120} \\
& {\bm R}_p^{(m+1)} = \Big[ {\bm \Delta}_1^{(m+1)} - {\bm W}_1^{(m+1)}, \; \cdots , \;           
      {\bm \Delta}_M^{(m+1)} - {\bm W}_M^{(m+1)} \Big] \label{alg130} \\
& {\bm R}_d^{(m+1)} =  \rho^{(m)} \Big[  {\bm W}_1^{(m+1)} - {\bm W}_1^{(m)}, \; \cdots , \;
   {\bm W}_M^{(m+1)} - {\bm W}_M^{(m)} \Big] \label{alg140} \\
&    \tau_{pri} =  p \sqrt{M} \, \tau_{abs} + \tau_{rel} \, \max ( e_1, e_2 ) \label{alg150} \\
&   \tau_{dual} =  p \sqrt{M} \, \tau_{abs} + \tau_{rel} \,  e_3 / \rho^{(m)} \, . \label{alg160}
\end{align}

Our overall ADMM-based optimization algorithm is as follows.
\begin{itemize}
\item[1.] Given $M$ and $K=2m_t+1$, calculate $\hat{\bm S}_{xk}$ and $\hat{\bm S}_{yk}$, $k \in [M]$ (computed using (\ref{specx}) and (\ref{specy})). Initialize iteration $\tilde{m}=0$,  $\tilde{\bm \Delta}^{(0)} = {\bm 0}$, 
$\bar{\tilde{\bm \Delta}} =  [ \bar{\bm \Delta}_1 , \; \bar{\bm \Delta}_2 , \; \cdots , \; \bar{\bm \Delta}_M] = \tilde{\bm \Delta}^{(0)}$ and use $\bar{\tilde{\bm \Delta}}$ to compute $\lambda_{ij}$'s.
\item[2.] Execute Algorithm \ref{alg0} with initial guess $\bar{\tilde{\bm \Delta}}$. Denote the resulting estimate by $\hat{\tilde{\bm \Delta}}$. Let $\tilde{m} \leftarrow \tilde{m}+1$.
\item[3.] Quit if using lasso, else set $\tilde{\bm \Delta}^{(\tilde{m})} = \hat{\tilde{\bm \Delta}}$ and $\bar{\tilde{\bm \Delta}} = \tilde{\bm \Delta}^{(\tilde{m})}$ to re-compute $\lambda_{ij}$'s via the LLA. 
\item[4.] Repeat steps 2 and 3 until convergence. 
\end{itemize}
A pseudocode for the above ADMM algorithm is given in Algorithm \ref{alg1}. It utilizes Algorithm \ref{alg0} in each LLA step.

For the numerical results in Secs.\ \ref{NE} and \ref{NEreal}, we used $\bar{\mu}=10$, $\bar{\rho}=2$, $\epsilon = 0.001$ for log-sum penalty, $a$=3.7 (as in \cite{Fan2001, Lam2009}) for the SCAD penalty, $\tau_{abs} = \tau_{rel} = 10^{-4}$, $m_{\max} = 200$, and $\tilde{m}_{\max} =1$ for lasso and $=2$ for LSP and SCAD penalties.
\vspace*{-0.04in}

\begin{algorithm} [htbp]
\caption{LLA-based ADMM Algorithm for Optimizing (\ref{deqn210})}
\label{alg1}
\algorithmicrequire{\; PSD estimators $\hat{\bm S}_{xk}$ and $\hat{\bm S}_{yk}$, $k \in [M]$ (computed using (\ref{specx}) and (\ref{specy})), regularization and penalty parameters $\lambda$ and $\rho=\bar{\rho}$, tolerances $\tau_{abs}$ and $\tau_{rel}$, variable penalty factor $\bar{\mu}$, maximum number of iterations $\tilde{m}_{\max}$}. For lasso penalty, $\tilde{m}_{\max} =1$ \\
\algorithmicensure{\;\ Estimated $\hat{\bm \Delta}_k$, $k \in [M]$, and edge-set $\hat{\cal E}_{\Delta}$}
\begin{algorithmic}[1] 
\STATE Initialize $\tilde{\bm \Delta}^{(0)} = {\bm 0}$ and 
$\bar{\tilde{\bm \Delta}} =  [ \bar{\bm \Delta}_1 , \; \bar{\bm \Delta}_2 , \; \cdots , \; \bar{\bm \Delta}_M] = \tilde{\bm \Delta}^{(0)} \in \mathbb{C}^{p \times pM}$. Set $\lambda_{ij} = \lambda$, $i,j \in [p]$. 
\STATE $\tilde{m}=1$
\WHILE{$\tilde{m} \le \tilde{m}_{\max}$,}
\STATE Execute Algorithm \ref{alg0}, resulting in output $\hat{\tilde{\bm \Delta}}$. Set $\tilde{\bm \Delta}^{(\tilde{m})} = \hat{\tilde{\bm \Delta}}$.
\IF{LSP/SCAD}
\STATE Set $\bar{\tilde{\bm \Delta}} = \tilde{\bm \Delta}^{(\tilde{m})}$ and re-compute $\lambda_{ij}$'s via the LLA (\ref{logsumInit}) or (\ref{scadInit}).
\ENDIF
\STATE $\tilde{m} \leftarrow \tilde{m}+1$
\ENDWHILE
\STATE With ${\bm \Delta}_k$, $k \in [M]$, denoting the converged estimates, set $\hat{\bm \Delta}_k = ({\bm \Delta}_k+{\bm \Delta}_k^H)/2$, $k \in [M]$, and 
\[ 
   \hat{\tilde{\bm \Delta}} = [ \hat{\bm \Delta}_1 , \; \hat{\bm \Delta}_2 , \; \cdots , \; \hat{\bm \Delta}_M] \,  .
\] 
If $\|\hat{\tilde{\bm \Delta}}^{(ij)} \| > 0$  assign edge $\{ i,j\} \in \hat{\cal E}_\Delta$, else $\{ i,j\} \not\in \hat{\cal E}_\Delta$. 
\end{algorithmic}
\end{algorithm}
\vspace*{-0.3in}

\subsection{Convergence} \label{conv2} The LLA-based objective function $\tilde{L}_f(\tilde{\bm \Delta})$, given by (\ref{admm}), is convex in $\tilde{\bm \Delta}$ (cf.\ Sec.\ \ref{DTL}). It is also closed, proper and lower semi-continuous. Hence, for any fixed $\rho > 0$, the ADMM algorithm is guaranteed to converge \cite[Sec.\ 3.2 and Appendix A]{Boyd2010}, in the sense that we have primal residual (\ref{alg130}) convergence to 0, dual residual (\ref{alg140}) convergence to 0, and the objective function $\tilde{L}_f(\tilde{\bm \Delta})$ convergence to the optimal value.

\subsection{BIC for Tuning Parameter Selection} \label{BIC}
Given $n$ and the chosen $K$ and $M$, for model selection we follow a BIC-like criterion similar to as given in \cite[Sec.\ III-E]{Tugnait2024} (which follows \cite{Yuan2017} who invokes \cite{Zhao2014}) for time-domain approaches. Let $| {\bm A} |_0$ denote the number of nonzero elements in ${\bm A}$ and suppose that $\hat{\tilde{\bm \Delta}} =  [ \hat{\bm \Delta}_1 , \; \hat{\bm \Delta}_2 , \; \cdots , \; \hat{\bm \Delta}_M]$ minimizes (\ref{deqn210}). We choose $\lambda$ to minimize $\mbox{BIC}(\lambda)$ given by
\begin{align} 
  \mbox{BIC}(\lambda) & = 4 K \sum_{k=1}^M \| \hat{\bm S}_{xk} \hat{\bm \Delta}_k \hat{\bm S}_{yk} - (\hat{\bm S}_{xk}
	   - \hat{\bm S}_{yk} ) \|_F \nonumber \\
		& \quad + \ln (4 K) \sum_{k=1}^M | \hat{\bm \Delta}_k |_0 \, .
  \label{eqn1800}  
\end{align}
Following \cite{Yuan2017} we use the term BIC (Bayesian information criterion) for it even though the cost function used is not negative log-likelihood although $\ln (4 K) \sum_{k=1}^M \| \hat{\bm \Delta}_k \|_0$ penalizes over-parametrization as in BIC. It is based on the fact that true ${\bm \Delta}_{k}^\diamond$ satisfies ${\bm S}_{xk}^\diamond {\bm \Delta}_{k}^\diamond {\bm S}_{yk}^\diamond - ({\bm S}_{xk}^\diamond- {\bm S}_{yk}^\diamond) = {\bm 0}$. Since (\ref{eqn1800}) is not scale invariant, we scale both $\hat{\bm S}_{xk}$ and $\hat{\bm S}_{yk}$ (and $\hat{\bm \Delta}_k$ commensurately) by $\bar{\bm \Sigma}^{-1}$ where $\bar{\bm \Sigma} = \mbox{diag}\{\hat{\bm \Sigma}_x\}$ is a diagonal matrix of diagonal elements of $\hat{\bm \Sigma}_x = \frac{1}{n_x} \sum_{t=1}^{n_x} {\bm x}(t) {\bm x}^\top(t) $ (we have $n_x=n_y=n$ in this paper). We have $M$ models, each with $K$ complex measurements ${\bm d}_x(\tilde{f}_{k,\ell})$   and ${\bm d}_y(\tilde{f}_{k,\ell}) $, leading to $4K$ real samples for each model: (\ref{eqn1800}) reflects that. 

In our numerical results we search over $\lambda \in [\lambda_{\ell} , \lambda_{u}]$,  where $\lambda_{\ell}$ and $\lambda_u$ are selected via a heuristic as in \cite{Tugnait2024}. Find the smallest $\lambda$, labeled $\lambda_{sm}$ for which we get a no-edge model; then we set $\lambda_{u}= \lambda_{sm}/2$ and $\lambda_{\ell} = \lambda_{u}/10$.

\section{Theoretical Analysis} \label{TA}  
Here we analyze some properties (consistency in inverse PSD difference estimation and graph recovery) of the minimizer of the convex function $\tilde{L}_f(\tilde{\bm \Delta})$ specified by (\ref{admm}), by following the approach of \cite{Negahban2012}. The approach of \cite{Negahban2012} requires $\lambda_{ij} >0$ for every $i,j \in p$, a condition that is violated by the SCAD penalty. Therefore, our theoretical analysis applies to the lasso and the log-sum penalties only. 

Define the true differential edgeset ${\cal E}_{\Delta}^\diamond$ and its cardinality $s$,
\begin{align} 
   {\cal E}_{\Delta}^\diamond  
	= & \Big\{ \{i,j\} ~:~ [({\bm S}_y^\diamond(f))^{-1} - ({\bm S}_y^\diamond(f))^{-1}]_{ij}
	     \not\equiv 0,  \nonumber \\
			& \quad\quad\quad   ~ 0 \le f \le 0.5 \Big\} \, , \quad
	    s = |{\cal E}_{\Delta}^\diamond| \, .
  \label{heqn200}  
\end{align}
In the rest of this section, we allow $p$, $K=2m_t+1$, $M$, $s$ and $\lambda$ to be a functions of sample size $n$, denoted as $p_n$, $K_n$, $M_n$, $s_n$ and $\lambda_n$, respectively. With $\tau > 2$, define
\begin{align}
  B_{xy} & = \max_{f \in [0,0.5]} \Big\{ \| {\bm S}_x^\diamond(f) \|_\infty \, , 
	           \| {\bm S}_y^\diamond(f) \|_\infty \Big\} \, , \label{heqn320}
\end{align}
\begin{align}
	B_d & = \max_{f \in [0,0.5]} \| ({\bm S}_y^\diamond(f))^{-1} 
	           - ({\bm S}_x^\diamond(f))^{-1} \|_\infty \, , \label{heqn321} 
\end{align}
\begin{align}
	\phi_{\min}^\diamond & = \min_{f \in [0,0.5]} \big\{ \phi_{\min}({\bm S}_x^\diamond(f))
	     \times \phi_{\min}({\bm S}_y^\diamond(f)) \big\},  \label{heqn322} 
\end{align}
\vspace*{-0.06in}
\begin{align}
\sigma_{xy} & = \max_{f \in [0,0.5], \ell \in [p]} \Big\{ [{\bm S}_x^\diamond(f) ]_{\ell \ell} \, , 
	           [{\bm S}_y^\diamond(f) ]_{\ell \ell} \Big\} \, ,\label{heqn323} 
\end{align}
\begin{align}
	C_0 & = 80 \, \sigma_{xy} \, \sqrt{2 \big(\ln(16 p_n^\tau M_n)/\ln(p_n) \big)} \, ,
	     \label{heqn324} 
\end{align}
\begin{align}
	N_1 & = \arg \min_n \Big\{ n ~:~ K_n > 2\, \ln(16 p_n^\tau M_n) \Big\} \, ,
	     \label{heqn325} 
\end{align}
\begin{align}
	N_2 & = \arg \min_n \Big\{ n ~:~ K_n > C_0^2\, \ln(p_n) /B_{xy} \Big\} \, , 
	     \label{heqn326} 
\end{align}
\begin{align}
	N_3 & = \arg \min_n \Big\{ n ~:~ \sqrt{K_n}/M_n  \ge \nonumber \\
	 & \quad\quad 768 \, B_{xy} B_{\rm init}^2 s_n C_0\, \sqrt{\ln(p_n)} / \phi_{\min}^\diamond \Big\} \, , 
	     \label{heqn326b} 
\end{align}
\begin{align}
	B_{\rm init} & = \left\{ \begin{array}{ll} 1 &  :  \mbox{lasso} \\
	           1+ \max_{i,j \in [p]} \| \bar{\tilde{\bm \Delta}}^{(ij)} \|/ \epsilon 
						 &  :  \mbox{log-sum}
						  \end{array} \right.  \label{heqn327}
\end{align}
where $\bar{\tilde{\bm \Delta}}$ is the initialization for LLA to the log-sum penalty (see (\ref{logsumInit})).

Let $\hat{\tilde{\bm \Delta}} = \arg \min_{\tilde{\bm \Delta}} \tilde{L}_f(\tilde{\bm \Delta})$ where $\tilde{L}_f(\tilde{\bm \Delta})$ is specified in (\ref{admm}). The proof of Theorem 1 is given in the Appendix \ref{append2}. \\
{\it Theorem 1 }: Under assumptions (A1)-(A2), if
\begin{align}  
 &  \lambda_n \ge  2 B_{\rm init} \sqrt{M_n} \big(6 \, B_{xy} B_d s_n  + 4 \big)
		     C_0 \sqrt{\frac{\ln(p_n)}{K_n}}  \, ,  \label{eqn450} 
\end{align}
\begin{align}
& n \ge   	\max \{N_1, N_2, N_3 \}\, ,
		     \label{eqn452}
\end{align}
then with probability $> 1- 2/p_n^{\tau -2}$, we have
\begin{align}
  & \| \hat{\tilde{\bm \Delta}} - \tilde{\bm \Delta}^\diamond \|_F 
	 = \sqrt{\sum_{k=1}^{M_n} \|\hat{\bm \Delta}_k - {\bm \Delta}_k^\diamond \|_F^2 } \le 
	 \frac{4 \sqrt{s_n} \,  \lambda_n}{\phi_{\min}^\diamond}
	       \label{eqn454}
\end{align}
for any $\tau >2$. 

{\it Remark 1}. {\it Convergence Rate}. If $B_{xy}$, $\sigma_{xy}$, $\phi_{\min}^\diamond$ and $B_d$ stay bounded with increasing sample size $n$, we have $\| \hat{\tilde{\bm \Delta}} - \tilde{\bm \Delta}^\diamond \|_F = {\cal O}_P (s_n^{1.5} \sqrt{ M_n \ln(p_n)/K_n})$. Therefore, for $\| \hat{\tilde{\bm \Delta}} - \tilde{\bm \Delta}^\diamond \|_F \rightarrow 0$ as $n \rightarrow \infty$, we must have $s_n^{1.5} \sqrt{ M_n \ln(p_n)/K_n}) \rightarrow 0$. Note that $K_n M_n \approx n/2$, therefore, for $\| \hat{\tilde{\bm \Delta}} - \tilde{\bm \Delta}^\diamond \|_F \rightarrow 0$ as $n \rightarrow \infty$, we need $s_n^{1.5} \sqrt{ n \ln(p_n)/K_n^2}) \rightarrow 0$. $\;\; \Box$

We now address graph recovery. We follow the proof technique of \cite[Theorem 10]{Zhao2022} in establishing  Theorem 2 whose proof is in the Appendix \ref{append2}. For some $\gamma_n >0$, define
\begin{align}  
    \hat{\cal E}_\Delta = & \left\{ \{i,j\} \, : \, \| \hat{\tilde{\bm \Delta}}^{(ij)} \| > \gamma_n > 0
		     \right\} \, , \label{neq4020} 
\end{align}
\begin{align}
		\tilde{\cal E}_\Delta^\diamond = &   \left\{ \{i,j\} \, : \, \|(\tilde{\bm \Delta}^\diamond)^{(ij)} \| >  0
		     \right\} \, , \label{neq4022} 
\end{align}
\begin{align}
       \bar{\sigma}_n = &   \frac{4 \sqrt{s_n} \,  \lambda_n}{\phi_{\min}^\diamond}   \, ,  \label{neq4022} \\
		\nu  = &  \min_{\{i,j\} \in {{\cal E}_\Delta^\diamond}} \,  \| \big( ({\bm S}_y^\diamond(f))^{-1} 
	           - ({\bm S}_x^\diamond(f))^{-1} \big)^{(ij)} \|  \, , \label{neq4024} 
\end{align}
\begin{align}
			N_4 = & \arg \min \Big\{ n \, : \, \bar{\sigma}_n \le 0.4 \nu \Big\} \, . \label{neq4025}
\end{align}
{\it Theorem 2}. For $\gamma_n = 0.5 \nu$ and $n \ge N_4$, $\hat{\cal E}_\Delta =\tilde{\cal E}_\Delta^\diamond$ with probability $> \; 1-2/p_n^{\tau-2}$ under the conditions of Theorem 1.  

\begin{figure*}[htbp]
\begin{center}
\includegraphics[width=0.8\linewidth]{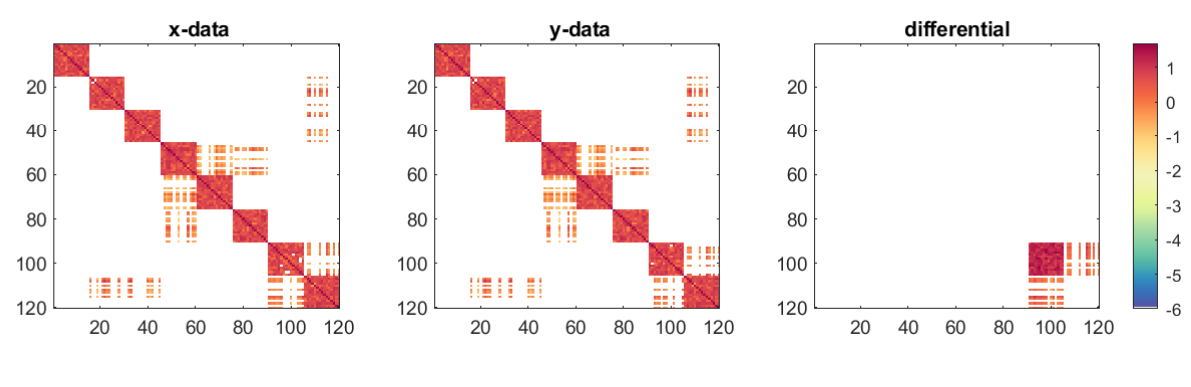}
\caption{\small{True $\log_{10} \big(\sum_{f=0:0.01:5} | [{\bm S}_x^\diamond(f)]_{ij} | \big)$ (left), $\log_{10} \big(\sum_{f=0:0.01:5} | [{\bm S}_y^\diamond(f)]_{ij} | \big)$ (middle), and $\log_{10} \big(\sum_{f=0:0.01:5} | [({\bm S}_y^\diamond(f))^{-1} - ({\bm S}_x^\diamond(f))^{-1}]_{ij} | \big)$ (right), $i,j \in [120]$, for the AR model, for a single Monte Carlo run: $p=120$ nodes.}} \label{figsyn1}
\end{center}
\end{figure*}

\begin{figure*}[htbp]
\begin{center}
\includegraphics[width=0.8\linewidth]{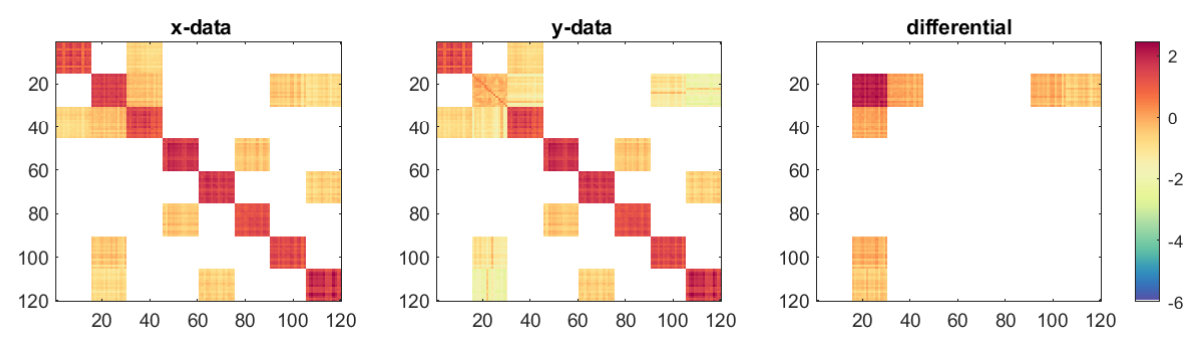}
\caption{\small{True $\log_{10} \big(\sum_{f=0:0.01:5} | [{\bm S}_x^\diamond(f)]_{ij} | \big)$ (left), $\log_{10} \big(\sum_{f=0:0.01:5} | [{\bm S}_y^\diamond(f)]_{ij} | \big)$ (middle), and $\log_{10} \big(\sum_{f=0:0.01:5} | [({\bm S}_y^\diamond(f))^{-1} - ({\bm S}_x^\diamond(f))^{-1}]_{ij} | \big)$ (right), $i,j \in [120]$, for the MA model, for a single Monte Carlo run: $p=120$ nodes.}} \label{figsyn2}
\end{center}
\end{figure*}

\section{Synthetic Data Examples} \label{NE}
We now present numerical results using synthetic data to illustrate the proposed approach (real data results are in Sec.\ \ref{NEreal}). In synthetic data examples the ground truth is known and this allows for assessment of the efficacy of various approaches in graph learning.

\subsection{Graphs with 120 nodes}  \label{NE1}
We consider two models for time-dependent data generation with $p=120$. 

\subsubsection{AR model} The time series data $\{ {\bm x}(t) \}$, ${\bm x}(t) \in \mathbb{R}^p$, is generated using a vector autoregressive (AR) model of order 3 (VAR(3)) as follows. Let $\{ {\bm w}(t) \}$, ${\bm w}(t) \in \mathbb{R}^p$, denote an i.i.d.\ zero-mean Gaussian sequence with precision matrix ${\bm \Omega}$ and let square matrices ${\bm A}_i \in \mathbb{R}^{p \times p}$, $i \in [3]$, be block-diagonal with $15 \times 15$ sub-blocks ${\bm A}^{(q)}_i$, $q \in [8]$. Then $\{ {\bm x}(t) \}$ is generated as
\begin{equation}  \label{VAR1}
    {\bm x}(t) = \sum_{i=1}^3 {\bm A}_i {\bm x}(t-i) + {\bm w}(t)  \, .
\end{equation} 
The diagonal entries of ${\bm \Omega}$ are set to 0.5, and the off-diagonal entries follow an Erd\"{o}s-R\`{e}nyi (ER) graph with connection probability $p_{er} = 0.001$: if nodes $j$ and $k$ are not connected in the ER graph, we have  $[\bm{\Omega}]_{jk} = 0$, and if they are connected, then $[\bm{\Omega}]_{jk}$ is uniformly distributed over $[-0.4,-0.1] \cup [0.1,0.4]$. Only 20\% of entries of ${\bm A}^{(q)}_i$'s are nonzero (randomly picked) and the nonzero elements are independently and uniformly distributed over $[-0.8,-0.3] \cup [0.3,0.8]$. We then check if the VAR(3) model is stable with all eigenvalues of the companion matrix $\le 0.95$ in magnitude; if not, the we scale ${\bm A}_i$'s to fulfill this condition (see \cite[Sec.\ VI.A]{Tugnait2025c}, \cite[Sec.\ 6.1]{Tugnait2024c}). To generate ${\bm y}$-data, we randomly eliminate one of the 8 clusters (${\bm A}^{(q)}_i$'s for randomly picked $q$) 
of ${\bm x}(t)$ and replace it with an independently generated ${\bm A}^{(q)}_i$, $i \in [3]$.

\subsubsection{MA model}  Here the time series data $\{ {\bm x}(t) \}$, ${\bm x}(t) \in \mathbb{R}^p$,  is generated using a vector moving average (MA) model of order 3 (MA(3)) as follows. Let $\{ {\bm w}(t) \}$, ${\bm w}(t) \in \mathbb{R}^p$, with precision matrix ${\bm \Omega}$, be as for the AR model, and let square matrices ${\bm B}_i \in \mathbb{R}^{p \times p}$, $i \in [3]$, be block-diagonal with $15 \times 15$ sub-blocks ${\bm B}^{(q)}_i$, $q \in [8]$. Then $\{ {\bm x}(t) \}$ is generated as
\begin{equation}  \label{VMA1}
    {\bm x}(t) = 0.5 {\bm I}_p {\bm w}(t) + \sum_{i=1}^3 ({\bm B}_i/i) {\bm w}(t-i) \, .
\end{equation} 
We pick ${\bm \Omega}$ as for the AR model. Only 25\% of entries of ${\bm B}^{(q)}_i$'s are nonzero  (randomly picked) and the nonzero elements are independently and uniformly distributed over $[-0.4,-0.2] \cup [0.2,0.4]$. To generate ${\bm y}$-data, we randomly eliminate one of the 8 clusters (${\bm B}^{(q)}_i$'s for randomly picked $q$) of ${\bm x}(t)$ and replace it with an independently generated ${\bm B}^{(q)}_i$, $i \in [3]$, with nonzero entries uniformly distributed over $[-0.2,0.2]$.

For both models, the first 100 samples are discarded to eliminate transients, and  we generate $n=n_x=n_y$ observations for ${\bm x}(t)$ and ${\bm y}(t)$, with $n \in \{512, 2048, 4096\}$. In each run, we calculate the true PSDs ${\bm S}_x^\diamond(f)$ and ${\bm S}_y^\diamond(f)$ for $f \in [0,0.5]$ at intervals of 0.01. Let $F$ denote the number of frequencies points in $[0,0.5]$ at intervals of 0.01. Define ${\bm \Delta}^\diamond(f) = ({\bm S}_y^\diamond(f))^{-1} - ({\bm S}_x^\diamond(f))^{-1} $, $b = \max_{i,j \in [p]} (1/F) \sum_f |[({\bm S}_x^\diamond(f))^{-1}]_{ij}|$, and  $d_{ij} = (1/F) \sum_f |[{\bm \Delta}^\diamond(f)]_{ij}|$. In each run, we take $\{i,j \} \in {\cal E}_\Delta^\diamond$ if $d_{ij} > \tau b$, else $\{i,j \} \not\in {\cal E}_\Delta^\diamond$, where the threshold $\tau = 0.001$ for the MA model and $=0.01$ for the AR model. To avoid very ``peaky'' inverse PSDs, if $b > 50,000$ we redraw the samples till this condition is satisfied: it is needed for the MA models. For a typical realization (run), Figs.\ \ref{figsyn1} and \ref{figsyn2} show heatmaps of $\log_{10} \big(\sum_{f=0:0.01:5} | [S^{-1}(f)]_{ij} | \big)$, $i,j \in [120]$, for the AR and MA models, respectively. For the chosen AR model, the percentage of distinct connected edges in the differential graph turn out to be $2.0 \pm 0.4$\% and for the MA model, they are $2.0 \pm 1.0$\%.

Simulation results based on 100 runs are shown in Table \ref{table1} where the performance measures are $F_1$-score and Hamming distance (between the estimated and true edgesets) for efficacy in edge detection, and timing per run as a surrogate for computational complexity. All algorithms were run on a Window 11 Enterprise operating system with processor Intel(R) Core(TM) i7-10700 CPU @2.90 GHz with 32 GB RAM, using MATLAB R2023a. We implemented our three proposed approaches, labeled ``DTS-FD, log-sum'', ``DTS-FD, lasso'' and  ``DTS-FD, SCAD''  (DTS stands for dependent time series and FD stands for frequency-domain) using log-sum, lasso and SCAD penalties, respectively. For comparison, we implemented two approaches that assume the data is i.i.d.\ and they are time-domain approaches based on sample covariances. One of them is based on \cite{Jiang2018} which minimizes lasso-penalized (\ref{eqn15}) based on difference of precision matrices (labeled ``IID, lasso'') and the other follows the recent approach of \cite{Tugnait2025a} (labeled ``IID, log-sum'') and it minimizes log-sum-penalized (\ref{eqn15}) based on difference of precision matrices. Although the approach of \cite{Tugnait2025a} is aimed at multi-attribute graphs, the approach therein applies to our problem by setting the number of attribute to one. For our proposed frequency-domain approaches, we used $M=2,4,5$ ($K=127, 255, 409$) for the MA model and $M=2,4,6$ ($K=127, 255, 341$) for the AR model, for $n=512, 2048, 4096$, respectively. 

\begin{table*}[htbp]
\caption{\small{\it $F_1$ scores, Hamming distances and timings for the synthetic data examples ($p=120$), averaged over 100 runs (standard deviation $\sigma$ in parentheses). ``DTS-FD, log-sum'', ``DTS-FD, lasso'' and ``DTS-FD, SCAD'' are the proposed approaches with log-sum, lasso and SCAD penalties, respectively, ``IID, lasso'' is the time-domain approach of \cite{Jiang2018} (also \cite{Yuan2017}) with lasso penalty, and ``IID, log-sum'' is the time-domain approach of \cite{Tugnait2025a} with log-sum penalty.}} \label{table1} 
\vspace*{-0.15in}
\begin{center}
\begin{tabular}{c|ccc|ccc}   \hline\hline
 $n$ &  512 &  2048  & 4096  &  512 &  2048  & 4096  \\  \hline\hline
\multicolumn{7}{c}{ $\lambda$'s picked to maximize $F_1$ score } \\
\multicolumn{4}{c|}{MA model: $F_1$ score ($\sigma$) } 
 & \multicolumn{3}{c}{AR model: $F_1$ score ($\sigma$)}\\ \hline
IID, lasso \cite{Jiang2018}  &  0.21 (0.06)  &  0.32 (0.09)  & 0.43 (0.10)   
  & 0.25 (0.09)  &  0.46 (0.10)  & 0.54 (0.10) \\
IID, log-sum \cite{Tugnait2025a}  &  0.28 (0.06)  &  0.46 (0.10)  & 0.58 (0.11)   
  & {0.32 (0.07)}  &  {0.53 (0.07)}  & {0.62 (0.09)} \\
DTS-FD, lasso   &  0.25 (0.09)  &  0.48 (0.20)    & 0.63 (0.19)   
  & 0.40 (0.12) & 0.65 (0.11) & 0.69 (0.12) \\ 
DTS-FD, log-sum   &  0.46 (0.10)  &  0.81 (0.17)  & 0.91 (0.15)   
  & 0.54 (0.10) & 0.79 (0.09) & 0.82 (0.09) \\ 
DTS-FD, SCAD   &  0.24 (0.08)  &  0.48 (0.20)  & 0.68 (0.19)   
  & 0.41 (0.12) & 0.65 (0.11) & 0.69 (0.12) \\ \hline\hline
	\multicolumn{4}{c|}{MA model: Hamming distance ($\sigma$) } 
 & \multicolumn{3}{c}{AR model: Hamming distance ($\sigma$)}\\ \hline
IID, lasso   &  232.2 (159.4)  &  220.0 (162.8)  & 143.0 (97.9)   
  & 191.6 (107.0) & 157.8 (101.3) & 132.1 (80.6) \\
IID, log-sum   &  173.2 (62.4)  &  136.2 (85.1)  & 105.2 (78.9)  
  & 153.3 (32.7) & 112.6 (28.2) & 94.0 (38.1) \\
DTS-FD, lasso   &  373.5 (361.5)  &  322.2 (392.6)  & 146.4 (191.1)   
  & 203.1 (172.4) & 100.0 (76.1) & 98.7 (85.4) \\ 
DTS-FD, log-sum   &  128.9 (69.3)  &  69.4 (88.1)  & 37.2 (79.1)   
  & 125.0 (79.0) & 55.6 (31.8) & 50.4 (36.3) \\ 
DTS-FD, SCAD   &  386.3 (372.6)  &  330.5 (408.6)  & 139.9 (189.5)   
  & 203.4 (172.5) & 100.2 (75.9) & 99.0 (85.3) \\ \hline\hline
	\multicolumn{4}{c|}{MA model: Timing (s) ($\sigma$) } 
 & \multicolumn{3}{c}{AR model: Timing (s)  ($\sigma$)}\\ \hline
IID, lasso  &  0.013 (0.003)  &  0.010 (0.002)  & 0.011 (0.002)   
  & 0.010 (0.004)  &  0.008 (0.003)  & 0.009 (0.003) \\
IID, log-sum  &  0.040 (0.005)  & 0.032 (0.007)  & 0.030 (0.005)   
  & 0.034 (0.007)  &  0.028 (0.004)  & 0.026 (0.003) \\
DTS-FD, lasso   &  7.6 (0.6)  &  8.9 (1.4)    & 11.7 (3.1)   
  & 12.6 (3.9) & 16.1 (4.2) & 17.5 (3.8) \\ 
DTS-FD, log-sum   &  17.4 (2.5)  &  23.9 (4.0)  & 33.4 (6.3)   
  & 21.2 (5.7) & 24.6 (6.2) & 28.3 (6.7) \\ 
DTS-FD, SCAD   &  16.5 (1.5)  &  22.2 (1.9)  & 26.8 (6.6)   
  & 26.2 (7.8) & 33.8 (8.9) & 35.6 (7.6) \\ \hline\hline
	\multicolumn{7}{c}{ $\lambda$'s picked to minimize BIC } \\
\multicolumn{4}{c|}{MA model: $F_1$ score ($\sigma$) } 
 & \multicolumn{3}{c}{AR model: $F_1$ score ($\sigma$)}\\ \hline
DTS-FD, log-sum  &  0.47 (0.11) & 0.78 (0.17) & 0.86 (0.15)   
  & 0.51 (0.12) & 0.75 (0.10) & 0.79 (0.10) \\ \hline\hline
	\multicolumn{4}{c|}{MA model: Hamming distance ($\sigma$) } 
 & \multicolumn{3}{c}{AR model: Hamming distance ($\sigma$)}\\ \hline
DTS-FD, log-sum   &  160.4 (97.6) & 69.9 (85.6) & 48.4 (78.3)   
  & 183.7 (163.1) & 60.9 (30.9) & 57.1 (36.8)  \\  \hline\hline
\end{tabular} 

\end{center}
\end{table*}

It is seen from Table \ref{table1} that our log-sum-penalized graph estimator significantly outperforms our lasso based graph estimator which in turn, significantly outperforms  the lasso based methods of \cite{Yuan2017, Jiang2018}, yielding higher $F_1$ scores and lower Hamming distances (ideal $F_1$ score is 1 and ideal Hamming distance is 0). The performance of our SCAD-penalized graph estimator in Table \ref{table1} is similar to that of our lasso based graph estimator showing little improvement. While the log-sum penalized ``IID, log-sum'' method of \cite{Tugnait2025a} improves upon ``IID, lasso'', it is significantly inferior to our proposed ``DTS-FD: log-sum''.

The improvement in performance with log-sum penalty over lasso (e.g., ``DTS-FD, log-sum'' over ``DTS-FD, lasso''), and  with dependent time-series (DTS-FD) modeling over i.i.d.\ data modeling (IID) approaches, comes at the cost of much increased computational time. For the MA model, we see from Table \ref{table1} that for sample sizes of $n=512$ and 4096, ``DTS-FD, log-sum'' approach yields $F_1$ scores of 0.46 and 0.91, respectively, compared to the ``DTS-FD, lasso'' $F_1$  scores of 0.28 and 0.58, respectively, which represent improvements by 84\% and 44\% (log-sum over lasso), respectively. But the computational cost (timing per run) increases (log-sum over lasso) by factors of 2.29 and 2.28 for $n=512$ and 4096, respectively. Note that the log-sum solution uses the lasso solution as an initial guess for LLA, and the lasso timing is included in the log-sum timing. One would expect the log-sum solution timing to be approximately twice the lasso timing: solve with lasso, use lasso-based LLA to obtain $\lambda_{ij}$'s and solve again using ADMM. On the other hand, we see little improvement in the $F_1$ score over lasso with the SCAD penalty even though the computational cost for SCAD is comparable to that for log-sum penalty.

\begin{figure*}[htbp]
\begin{subfigure}[b]{0.5\textwidth}
\begin{center}
\includegraphics[width=0.9\linewidth]{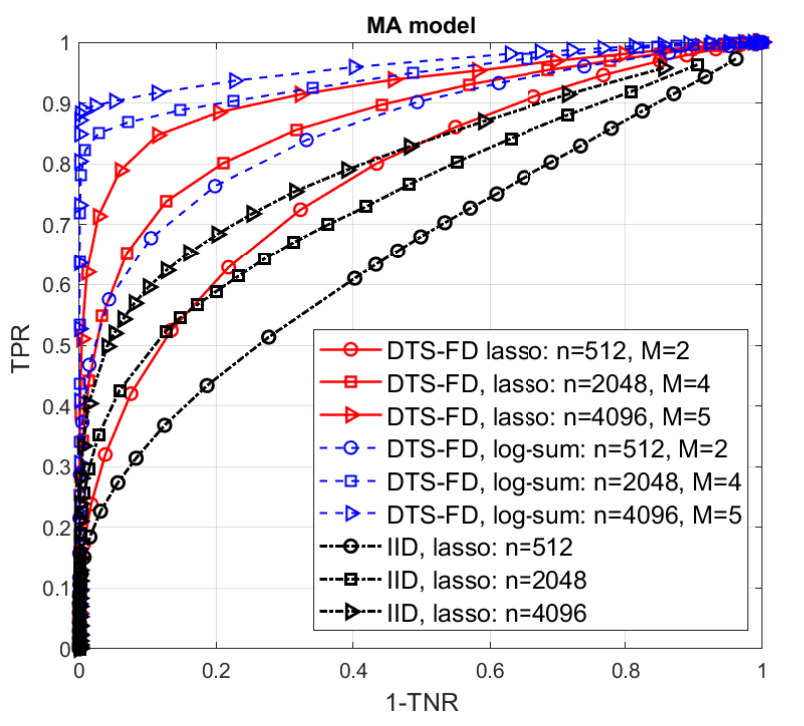}
\caption{ROC curves for the moving average (MA) model.}\label{figsynMA}
\end{center}
\end{subfigure}%
\begin{subfigure}[b]{0.5\textwidth}
\begin{center}
\includegraphics[width=0.9\linewidth]{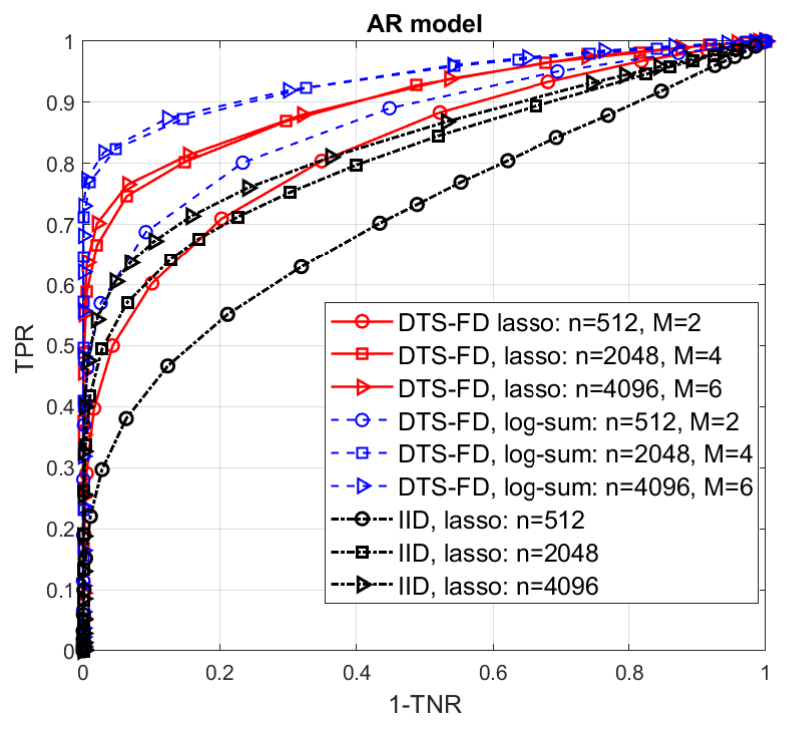}
\caption{ROC curves for the autoregressive (AR) model.}\label{figsynAR}
\end{center}
\end{subfigure}%
\caption{\small{ROC curves: ``DTS-FD, log-sum'' is the proposed approach with log-sum penalty, ``DTS-FD, lasso'' is the proposed approach with lasso penalty, and \textsc{``IID, lasso''} is the time-domain approach of \cite{Jiang2018} (also \cite{Yuan2017}) with lasso penalty. TPR=true positive rate, TNR=true negative rate.}} \label{figsyn}
\end{figure*}

For the AR model, we see smaller (compared to the MA model) yet significant improvements in the $F_1$ scores with log-sum penalty over lasso in Table \ref{table1}. For sample sizes of $n=512$ and 4096, ``DTS-FD, log-sum'' approach yields $F_1$ scores of 0.54 and 0.82, respectively, compared to the ``DTS-FD, lasso'' $F_1$  scores of 0.40 and 0.69, respectively, which represent improvements by 26\% and 19\% (log-sum over lasso), respectively. The timing per run increases (log-sum over lasso) by factors of 1.68 and 1.63 for $n=512$ and 4096, respectively. 

When analyzing the trade-off between performance and computational time for lasso and log-sum penalties, the Hamming distance performance measure seems to provide a ``better'' metric. The Hamming distance between the true and the estimated graph edgeset is the sum of the number of distinct incorrect edges in the estimated edgeset (a true edge is missing, or an edge missing from true edgeset is present in the estimated edgeset). For the MA model, for sample sizes of $n=512$ and 4096, ``DTS-FD, lasso'' approach yields Hamming distances of 373.5 and 146.4, respectively, compared to the ``DTS-FD, log-sum'' Hamming distances of 128.9 and 37.2, respectively, which represent reductions by factors of 2.90 and 3.93 (log-sum over lasso), respectively. For the AR model, for sample sizes of $n=512$ and 4096, ``DTS-FD, lasso'' approach yields Hamming distances of 203.1 and 98.7, respectively, compared to the ``DTS-FD, log-sum'' Hamming distances of 125.0 and 50.4, respectively, which represent reductions by factors of 1.62 and 1.96 (log-sum over lasso), respectively. Thus we have reduction in the Hamming distance by factors of 2.90 and 3.93 with a computational cost increase by factors of 2.29 and 2.28 for $n=512$ and 4096, respectively, for the MA model, and reduction in the Hamming distance by factors of 1.62 and 1.96 with a computational cost increase by factors of 1.68 and 1.63 for $n=512$ and 4096, respectively, for the AR model.   Since the main objective of differential graph learning is determination of the true edgeset, such a trade-off seems to be reasona

In Table \ref{table1}, $\lambda$'s were first picked from a grid of values to maximize the $F_1$ score (ground truth is known in synthetic data examples) -- this establishes how well a method will perform if $\lambda$'s are judiciously picked. For log-sum penalty we also show the results when $\lambda$'s are selected to minimize the BIC criterion of Sec.\ \ref{BIC}.   We see that the heuristic BIC-type criterion performs well.

\begin{table*}[htbp]
\caption{{\small{\it $F_1$ scores, Hamming distances and timings for AR(3) model with $p \in \{60,120,240\}$, averaged over 100 runs (standard deviation $\sigma$ in parentheses). ``DTS-FD, log-sum'' and ``DTS-FD, lasso'' are the proposed approaches with log-sum and lasso penalties, respectively. Also shown is the normalized Hamming distance which is the Hamming distance divided by total number of distinct edges in the differential graph, expressed as percentage.}}} \label{table2} 
\vspace*{-0.15in}
\begin{center}
\begin{tabular}{c|ccc|ccc}   \hline\hline
 $n$ &  512 &  2048  & 4096  &  512 &  2048  & 4096  \\  \hline\hline
\multicolumn{7}{c}{ AR model with varying $p$, $\lambda$'s picked to maximize $F_1$ score } \\
\multicolumn{4}{c|}{DTS-FD, lasso: $F_1$ score ($\sigma$) } 
 & \multicolumn{3}{c}{DTS-FD, log-sum: $F_1$ score ($\sigma$)}\\ \hline
$p$=60  &  0.42 (0.16)  &  0.64 (0.16)  & 0.74 (0.13)   
  & 0.65 (0.12)  &  0.85 (0.10)  & 0.89 (0.10) \\
$p$=120  & 0.40 (0.12) & 0.65 (0.11) & 0.69 (0.12)   
  & 0.54 (0.10) & 0.79 (0.09) & 0.82 (0.09) \\
$p$=240   &  0.39 (0.09)  &  0.61 (0.07)    & 0.68 (0.06)   
  & 0.40 (0.13) & 0.65 (0.07) & 0.73 (0.07) \\  \hline\hline
	\multicolumn{4}{c|}{DTS-FD, lasso: Hamming distance ($\sigma$) } 
 & \multicolumn{3}{c}{DTS-FD, log-sum: Hamming distance ($\sigma$)}\\ \hline
$p$=60    &  86.1 (75.2)  &  54.8 (48.6)  & 31.2 (19.5)   
  & 37.4 (15.4) & 18.7 (15.3) & 14.7 (15.3) \\
$p$=120  & 203.1 (172.4) & 100.0 (76.1) & 98.7 (85.4)   
  & 125.0 (79.0) & 55.6 (31.8) & 50.4 (36.3) \\ 
$p$=240  &  734.9 (63.9) & 364.7 (97.9) & 310.1 (106.6)   
  & 1464 (300.8) & 374.4 (212.4) & 295.5 (120.4) \\ \hline
	\multicolumn{4}{c|}{DTS-FD, lasso: normalized Hamming distance \% ($\sigma$) } 
 & \multicolumn{3}{c}{DTS-FD, log-sum: normalized Hamming distance \%  ($\sigma$)}\\ \hline
$p$=60    &  4.71 (4.11)  &  2.99 (2.66)  & 1.70 (1.07)   
  & 2.04 (0.84) & 1.02 (0.84) & 0.80 (0.84) \\
$p$=120  & 2.80 (2.37) & 1.38 (1.05) & 1.36 (1.18) 
  & 1.72 (1.09) & 0.77 (0.44) & 0.69 (0.50) \\ 
$p$=240  &  2.54 (0.22) & 1.26 (0.34) & 1.07 (0.37)   
  & 5.06  (1.04) & 1.29 (0.73) & 1.02 (0.42) \\ \hline\hline
	\multicolumn{4}{c|}{DTS-FD, lasso: Timing (s) ($\sigma$) } 
 & \multicolumn{3}{c}{DTS-FD, log-sum: Timing (s)  ($\sigma$)}\\ \hline
$p$=60    &  3.30 (1.37)  &  3.40 (0.93)  & 3.86 (1.03)   
  & 4.79 (1.69)  &  4.87 (1.13)  & 5.09 (1.10) \\
$p$=120    & 12.6 (3.9) & 16.1 (4.2) & 17.5 (3.8) 
  & 21.2 (5.7) & 24.6 (6.2) & 28.3 (6.7) \\ 
$p$=240    &  63.5 (16.1)  &  103.5 (23.8)  & 135.4 (27.7)   
  & 135.7(44.8) & 134.7 (31.2) & 142.8 (30.0) \\ \hline\hline
\end{tabular} 
\end{center}
\end{table*}

The receiver operating characteristic (ROC) curves are shown in Fig.\ \ref{figsyn} for three approaches ``DTS-FD, log-sum'', ``DTS-FD, lasso'' and ``IID, lasso''. By changing the penalty parameter $\lambda$ and determining the resulting edges over 100 runs, we calculated the true positive rate (TPR) which calculates true edges correctly detected ($\| \hat{\tilde{\bm \Delta}}^{(ij)} \| \ne 0$ and $\| (\tilde{\bm \Delta}^\diamond)^{(ij)} \| \ne 0$), and false positive rate 1-TNR (where TNR is the true negative rate) which are the edges $\{i,j\}$ for which $\| \hat{\tilde{\bm \Delta}}^{(ij)} \| \ne 0$ but  $\| (\tilde{\bm \Delta}^\diamond)^{(ij)} \| = 0$.  It is seen from Fig.\ \ref{figsyn} that our log-sum-penalized graph estimator significantly outperforms both the ``IID, lasso'' approach and our lasso based graph estimator, yielding much higher TPR for a given 1-TNR, consistent with the results of Table \ref{table1}.

\subsection{Graphs with varying number of nodes} 
We now consider AR(3) models for time-dependent data generation with varying number of graph nodes $p \in \{60, 120, 240 \}$. The objective is to empirically study the performance stability of the proposed solutions with varying model dimensions. The AR(3) model follows (\ref{VAR1}) where ${\bm A}_i \in \mathbb{R}^{p \times p}$, $i \in [3]$, is block-diagonal with 
\begin{itemize}
\item[(i)] six $10 \times 10$ sub-blocks ${\bm A}^{(q)}_i$, $q \in [6]$, when $p=60$,
\item[(ii)] eight $15 \times 15$ sub-blocks ${\bm A}^{(q)}_i$, $q \in [8]$, when $p=120$ (as in Sec.\ \ref{NE1}),
\item[(iii)] eight $30 \times 30$ sub-blocks ${\bm A}^{(q)}_i$, $q \in [6]$, when $p=240$.
\end{itemize}
All other details regarding generation of ${\bm \Omega}$, ${\bm A}^{(q)}_i$'s, $\{{\bm x}(t)\}$ and $\{{\bm y}(t)\}$ are exactly as before in Sec.\ \ref{NE1}. The percentage of distinct connected edges in the differential graphs turn out to be $3.0 \pm 1.0$\%, $2.0 \pm 0.4$\% and $2.0 \pm 0.2$\% for $p=60$, $p=120$ and $p=240$, respectively.

Simulation results based on 100 runs are shown in Table \ref{table2} for the proposed ``DTS-FD, lasso'' and ``DTS-FD, log-sum'' approaches where the performance measures, as in Table \ref{table1},  are the $F_1$-score, the Hamming distance and timing per run. The results for $p=120$ are as in Table \ref{table1}. Since the number of distinct connected edges in the differential graph vary with $p$, we also show the the normalized Hamming distance which is the Hamming distance divided by total number of distinct edges in the differential graph, expressed as percentage. As for Table \ref{table1}, we used $M=2,4,6$ ($K=127, 255, 341$) for all AR models, for $n=512, 2048, 4096$, respectively. The number of unknowns in ${\bm \Delta}_k$ is $p^2$, therefore, the number of unknowns being estimated is $M p^2$. It is seen in Table \ref{table2} that the $F_1$ score decreases and the Hamming distance increases (i.e., the performance deteriorates) with increasing dimension $p$ for the same sample size $n$ since the number of unknowns being estimated increases. The performance is stable with increasing $p$ as the performance improves with increasing $n$, and the deterioration in the performance measures with increasing $p$ for fixed $n$ is ``gradual.''

\begin{figure*}[htbp]
\begin{subfigure}[b]{0.5\textwidth}
\begin{center}
\includegraphics[width=0.68\linewidth]{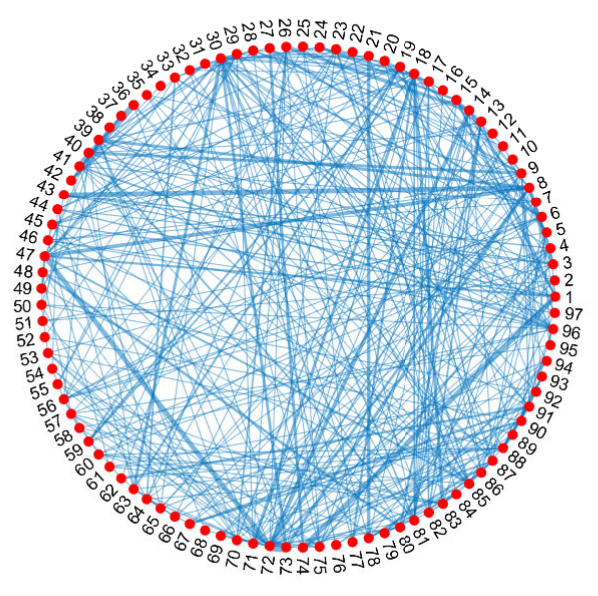}
\caption{IID, lasso: 429 edges}
\end{center}
\end{subfigure}%
\begin{subfigure}[b]{0.5\textwidth}
\begin{center}
\includegraphics[width=0.68\linewidth]{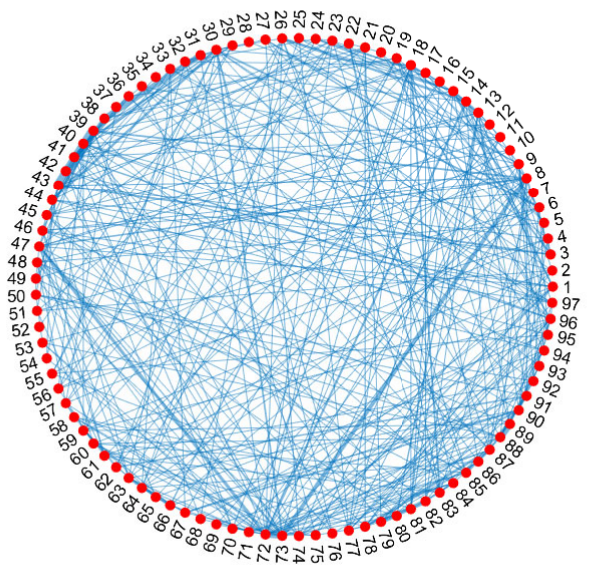}
\caption{IID, log-sum: 462 edges}
\end{center}
\end{subfigure}%
\newline
\begin{subfigure}[b]{0.5\textwidth}
\begin{center}
\includegraphics[width=0.68\linewidth]{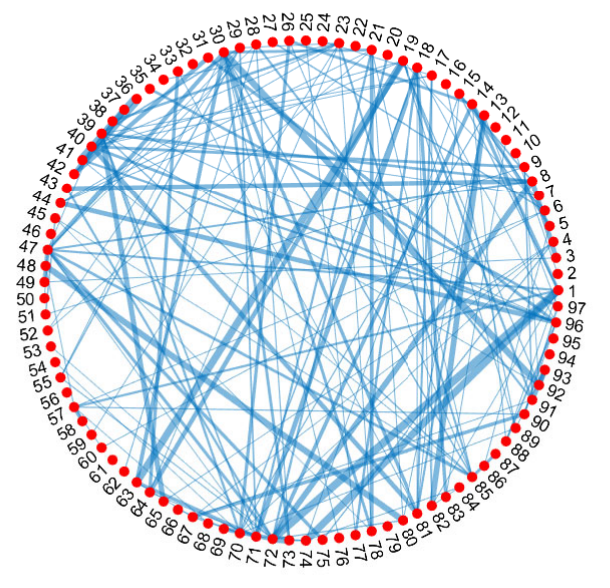}
\caption{Freq-domain, lasso: 205 edges}
\end{center}
\end{subfigure}%
\begin{subfigure}[b]{0.5\textwidth}
\begin{center}
\includegraphics[width=0.66\linewidth]{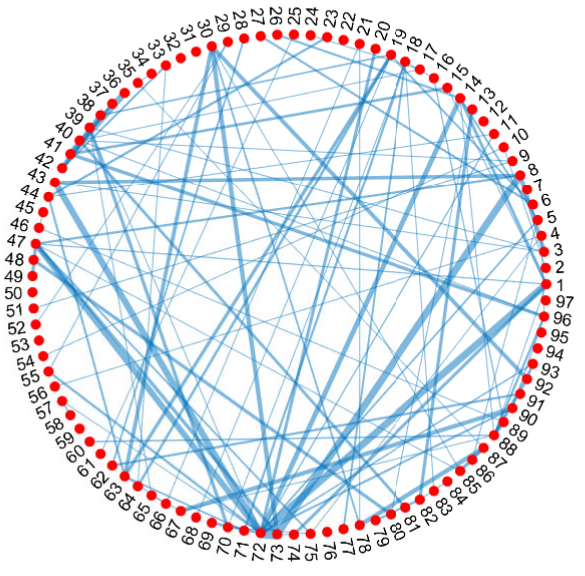}
\caption{Freq-domain, log-sum: 151 edges}
\end{center}
\end{subfigure}%
\caption{\small{Differential graphs comparing financial time series (S\&P 97 stocks share prices) over period Jan.\ 2, 2013 to Jan.\ 14, 2015 with that over period Dec.\ 17, 2015 to Jan.\ 1, 2018 (each series with 512 samples): (a) time-domain IID model with lasso penalty \cite{Jiang2018} (IID, lasso), (b) time-domain IID model with log-sum penalty \cite{Tugnait2025a} (IID, log-sum), (c) proposed freq-domain approach with group lasso penalty (FD-DTS, lasso), (d) proposed freq-domain approach with group log-sum penalty (FD-DTS, log-sum). In the freq-domain approaches we used $M=2$ ($m_t=63$, $K=127$). In the figures the thickness of the lines reflects the strength of the connection (determined by $\| \hat{\tilde{\bm \Delta}}^{(ij)} \|$).}} \label{figreal}
\end{figure*}

\section{Real Data: Financial Time Series} \label{NEreal}
Here we investigate differences in the time series graphical models of the share prices of 97 stocks in the S\&P 100 index over two different time periods: Jan.\ 2, 2013 to Jan.\ 14, 2015 and Dec.\ 17, 2015 to Jan.\ 1, 2018. In the real data example our goal is visualization and exploration of the differential conditional dependency structure underlying the data since the ground truth is unknown. The selection of the duration of each period leads to equal number of samples in the two time periods.

We consider daily share prices (at close of the day) of 97 stocks in the S\&P 100 index from Jan. 1, 2013 through Jan.\ 1, 2018. This data was gathered from Yahoo Finance website. If $z_m(t)$ is share price of $m$th stock on day $t$, we pre-process to create $x_m(t) = \ln (z_m(t)/z_m(t-1))$ as the time series to analyze. Such transformations are common in the analysis of financial time series. For instance, such pre-processing is used in \cite[Sec.\ 5.2]{Songsiri2010} for topology selection for graphical models for international stock market data, and in \cite[Sec.\ 5.2]{Chen2021} for analyzing GDP growth, total manufacturing production growth and consumer price index core inflation data. We have $x_m(t) = \ln(z_m(t)) - \ln(z_m(t-1))$ which implies that we first perform log($\cdot$) transformation (generally believed to make data ``more Gaussian''), followed by lag one differencing to make the data close to univariate uncorrelated and stationary. 

The 97 stocks in the S\&P 100 index are classified into 11 sectors (according to the Global Industry Classification Standard (GISC)) and we order the nodes to group them as information technology (nodes 1-12), health care (13-27), financials (28-44), real estate (45-46), consumer discretionary (47-56), industrials (57-68), communication services (69-76), consumer staples (77-87), energy (88-92), materials (93), utilities (94-97). For each $m$, $x_m(t)$ was centered and normalized to unit variance. The pre-processed data from Jan.\ 2, 2013 to Jan.\ 14, 2015 was taken as the ${\bm x}$-data and that from Dec.\ 17, 2015 to Jan.\ 1, 2018 (each series with 512 samples) was taken as the ${\bm y}$-data. The resulting differential graphs are shown in Fig.\ \ref{figreal}. The tuning parameter $\lambda$ as selected as discussed in Sec.\ \ref{BIC}. The proposed log-sum penalty yields the sparsest graph with 151 edges. Some of the ``strongly'' connected nodes (thicker lines and higher degrees) in Fig.\ \ref{figreal}(d) are Apple (labeled node 1), Meta (72), Alphabet (73), Microsoft (8), Visa (43), Amazon (47), Abbott Labs (14) and American Express (30). The IID model based differential graphs in Figs.\ \ref{figreal}(a) and \ref{figreal}(b) are just too dense.

{\section{Conclusion}} 
Estimation of  differences in CIGs of two TSGGMs was investigated where the two TSGGMs are known to have similar structure. We presented and analyzed a penalized D-trace loss function approach in the frequency domain for differential graph learning using both convex (group lasso) and non-convex (log-sum and SCAD group penalties) regularization functions. An ADMM algorithm was presented to optimize the objective function where, for non-convex penalties, a local linear approximation approach was used. A model selection method for tuning parameter selection was also presented. Both synthetic and real data examples were presented to illustrate the proposed approach where in synthetic data examples, our frequency-domain based log-sum-penalized differential time-series graph estimator significantly outperformed our frequency-domain based lasso-penalized differential time-series graph estimator, with $F_1$ score as the performance metric. Our frequency-domain estimators significantly outperformed the i.i.d.\ modeling based time domain methods of \cite{Yuan2017, Jiang2018} (lasso penalty) and \cite{Tugnait2025a} (log-sum penalty). The SCAD penalty resulted in little improvement over our lasso based graph estimator.

Theoretical analysis establishing sufficient conditions for consistency and graph recovery was presented using the framework of \cite{Negahban2012} which however, does not apply to the SCAD penalty. Exploration of alternative analysis techniques (e.g., \cite{Loh2017}) to handle penalties such as SCAD, and to analyze (\ref{deqn210}) instead of its LLA approximation, is of interest.

\appendices
\section{Derivation of (32)} \label{append1}
Now we derive (\ref{eqn120}). By (\ref{eqn116})-(\ref{eqn118}), we have
\begin{align} 
  {\bm I}_p \otimes {\bm I}_p  & = ({\bm Q}_y^\ast {\bm Q}_y^T) \otimes ({\bm Q}_x {\bm Q}_x^H)  \nonumber \\
	& = \big( {\bm Q}_y^\ast \otimes {\bm Q}_x \big) \big(  {\bm Q}_y^T \otimes {\bm Q}_x^H) \big) \nonumber \\
	& = \big( {\bm Q}_y^\ast \otimes {\bm Q}_x \big) \big( {\bm I}_p \otimes {\bm I}_p \big)
	     \big(  {\bm Q}_y^T \otimes {\bm Q}_x^H) \big) \label{ap1p10}
\end{align}
and
\begin{align} 
   \hat{\bm \Sigma}_y^\ast \otimes \hat{\bm \Sigma}_x & =
  \big( {\bm Q}_y^\ast \otimes {\bm Q}_x \big) \big( {\bm D}_y \otimes {\bm D}_x \big)
	     \big(  {\bm Q}_y^T \otimes {\bm Q}_x^H) \big) \, . \label{ap1p12}  
\end{align}
Let 
\begin{align}
 {\bm C} & = \hat{\bm \Sigma}_x-\hat{\bm \Sigma}_y + \frac{\rho}{2} ({\bm W}^{(i)} - {\bm U}^{(i)}) \, . \label{ap1p13}
\end{align}
Then by (\ref{eqn116}) and (\ref{eqn121}),
\begin{align} 
  {\rm vec}({\bm \Delta}) & = \big(  {\bm Q}_y^\top \otimes {\bm Q}_x^H)^{-1} \,
	    \big( {\bm D}_y \otimes {\bm D}_x + \frac{\rho}{2} {\bm I}_p \otimes {\bm I}_p \big)^{-1} \nonumber \\
	& \quad\quad \times \big( {\bm Q}_y^\ast \otimes {\bm Q}_x \big)^{-1} {\rm vec}({\bm C}) \nonumber 
\end{align}
\begin{align}
	& = \big( {\bm Q}_y^\ast \otimes {\bm Q}_x \big) \,
	    \big( {\bm D}_y \otimes {\bm D}_x + \frac{\rho}{2} {\bm I}_p \otimes {\bm I}_p \big)^{-1} \nonumber \\
	& \quad\quad \times \big( {\bm Q}_y^\top \otimes {\bm Q}_x^H \big) {\rm vec}({\bm C}) \nonumber 
\end{align}
\begin{align}
	&  = \big( {\bm Q}_y^\ast \otimes {\bm Q}_x \big) \,
	    \big( {\bm D}_y \otimes {\bm D}_x + \frac{\rho}{2} {\bm I}_p \otimes {\bm I}_p \big)^{-1} \nonumber \\
	& \quad\quad \times 
			 {\rm vec}({\bm Q}_x^H {\bm C} {\bm Q}_y)    \nonumber 
\end{align}
\begin{align}
	&  = \big( {\bm Q}_y^\ast \otimes {\bm Q}_x \big) \,
			 {\rm vec} \big({\bm D} \circ ({\bm Q}_x^H {\bm C} {\bm Q}_y) \big) \nonumber \\
	&  = {\rm vec} \big( {\bm Q}_x \left[ {\bm D} \circ ({\bm Q}_x^H {\bm C} {\bm Q}_y) \right] {\bm Q}_y^H \big) 
	\, . \label{ap1p14}  
\end{align}
The desired (\ref{eqn120}) follows from (\ref{ap1p13}) and (\ref{ap1p14}).

\section{. ~~Technical Lemmas and Proofs of Theorems 1 and 2} \label{append2}
For theoretical analysis we will use the restricted strong convexity (RSC) based results from \cite{Negahban2012} which are given therein for real-valued vectors variables. Therefore, we first express our cost $\tilde{L}_f(\tilde{\bm \Delta})$ in terms of ${\rm vec}({\bm \Delta}_k)$, $k \in [M_n]$, and then in terms of ${\rm vec}({\rm Re}({\bm \Delta}_k))$ and ${\rm vec}({\rm Im}({\bm \Delta}_k))$, before invoking \cite{Negahban2012}.

With ${\bm \psi}_k :=  {\rm vec}({\bm \Delta}_k)$ define 
\begin{align}
  {\bm \theta}_k := & \begin{bmatrix} {\rm Re}({\bm \psi}_k) \\
	    {\rm Im}({\bm \psi}_k) \end{bmatrix} \in \mathbb{R}^{2 p^2} \, , \quad
	\bar{\bm \psi}_k := & \begin{bmatrix} {\bm \psi}_k \\
	    {\bm \psi}_k^\ast \end{bmatrix} \in \mathbb{C}^{2 p^2} \, . \label{ap2ad1}
\end{align}
Then cost $\tilde{L}(\tilde{\bm \Delta})$ of (\ref{app1eq1000}) can be re-expressed in terms of $\bar{\bm \psi}_k$s and ${\bm \theta}_k$s as
\begin{align}
  {\cal L}_c(\tilde{\bm \psi}) & = \sum_{k=1}^{M_n}  \Big(
	  \frac{1}{2} \bar{\bm \psi}_k^H \bar{\cal H}_k \bar{\bm \psi}_k
		 - \bar{\bm \psi}_k^H \bar{\bm b}_k  \Big)   \label{ap2ad2} 
\end{align}
where
\begin{align}
  \tilde{\bm \psi} & := \begin{bmatrix} \bar{\bm \psi}_1^\top & \cdots & \bar{\bm \psi}_{M_n}
		\end{bmatrix}^\top  \in \mathbb{C}^{2p^2{M_n}} \, , \label{ap2ad3} 
\end{align}
\begin{align}
	\bar{\cal H}_k & := \begin{bmatrix} \hat{\bm S}_{yk}^\ast \otimes \hat{\bm S}_{xk}  & {\bm 0} \\
	           {\bm 0} & \hat{\bm S}_{yk} \otimes \hat{\bm S}_{xk}^\ast \end{bmatrix} 
			\in \mathbb{C}^{2p^2{M_n} \times 2p^2{M_n}}			 \, , \label{ap2ad4} 
\end{align}
\begin{align}
		\bar{\bm b}_k & = \begin{bmatrix} \mbox{vec}(\hat{\bm S}_{xk}-\hat{\bm S}_{yk}) \\ 
					 \mbox{vec}((\hat{\bm S}_{xk} - \hat{\bm S}_{yk})^\ast) \end{bmatrix}
				\in \mathbb{C}^{2p^2{M_n}}	    \, , \label{ap2ad5} 
\end{align}
and
\begin{align}
  {\cal L}_r(\tilde{\bm \theta}) & = \sum_{k=1}^{M_n}  \Big(
	  {\bm \theta}_k^H {\cal H}_k {\bm \theta}_k
		 - 2\, {\bm \theta}_k^\top {\bm b}_k  \Big)   \label{ap2ad6} 
\end{align}
where ($\iota = \sqrt{-1} \,$)
\begin{align}
  \tilde{\bm \theta} & := \begin{bmatrix} {\bm \theta}_1^\top & \cdots & {\bm \theta}_{M_n}^\top
		\end{bmatrix}^\top  \in \mathbb{R}^{2p^2M_n} \, , \label{ap2ad7} 
\end{align}
\begin{align}
	{\cal H}_k & := \frac{1}{2} {\bm T}_{rc}^H  \bar{\cal H}_k {\bm T}_{rc}
			\in \mathbb{R}^{2p^2M_n \times 2p^2M_n}			 \, ,  \label{ap2ad8a} 
\end{align}
\begin{align}
		{\bm b}_k & := \frac{1}{2} {\bm T}_{rc}^H \bar{\bm b}_k 	\in \mathbb{R}^{2p^2M_n} 
		\, , \quad   \tilde{\bm T}_{rc}  := \begin{bmatrix} 1 & \iota \\ 1 & -\iota  \end{bmatrix} \, , \label{ap2ad8} 
\end{align}
\begin{align}
		{\bm T}_{rc} & := \tilde{\bm T}_{rc} \otimes {\bm I}_{p^2}
				\in \mathbb{C}^{2p^2}	    \, , \quad \bar{\bm \psi}_k = {\bm T}_{rc} {\bm \theta}_k \, ,
		\label{ap2ad9} 
\end{align}
and ${\bm T}_{rc}$ yields real-to-complex transformation \cite[Appendix 2]{Schreier10}. Note that we have the equalities $\tilde{L}(\tilde{\bm \Delta}) = {\cal L}_c(\tilde{\bm \psi}) = {\cal L}_r(\tilde{\bm \theta})$. It is easy to establish that $\|\tilde{\bm T}_{rc}\| = \|{\bm T}_{rc}\| = \sqrt{2}$ and $\|\tilde{\bm T}_{rc}\|_{1,\infty} = \|{\bm T}_{rc}\|_{1,\infty} = \|{\bm T}_{rc}^H\|_{1,\infty} = \|\tilde{\bm T}_{rc}^H\|_{1,\infty} =2$.

We now turn our attention to the penalty/regularization term $\sum_{i, j=1}^p \lambda_{ij}  \| \tilde{\bm \Delta}^{(ij)} \| $ in (\ref{admm}) and will express it to conform to the framework of \cite{Negahban2012}. Note that the term $\tilde{\bm \Delta}^{(ij)}$ corresponds to the edge $\{i,j\}$ of the graph. We denote its real-valued version as 
\begin{equation}
	\tilde{\bm \theta}_{Gt} = \big[{\rm Re}(\tilde{\bm \Delta}^{(ij)})^\top \;\; 
	     {\rm Im}(\tilde{\bm \Delta}^{(ij)})^\top \big]^\top \in \mathbb{R}^{2M_n}
	\label{ap2ad10}
\end{equation}
(subscript $G$ for grouped variables \cite{Negahban2012}), with index $t \in [p^2]$, $(i,j) \leftrightarrow t = (i-1)p+j$ and $i=\lfloor t/p \rfloor +1$, $j =t \, \bmod \, p$. Using this notation, we have (we now denote $\lambda$ by $\lambda_n$)
\begin{align}
&	\sum_{i, j=1}^p \lambda_{ij}  \| \tilde{\bm \Delta}^{(ij)} \| 
	  = \lambda_n \sum_{t=1}^{p^2} w_t \| \tilde{\bm \theta}_{Gt} \|_2  \, , \label{ap2ad12} \\
&	w_t  = \left\{ \begin{array}{ll} 1 &  :  \mbox{lasso} \\
	         \epsilon  / (\epsilon + \|\bar{\tilde{\bm \theta}}_{Gt} \|)
						 &  :  \mbox{log-sum} \, ,
						  \end{array} \right.  \label{ap2ad13}
\end{align}
where $\bar{\tilde{\bm \theta}}_{Gt}$ corresponds to $\bar{\tilde{\bm \Delta}}^{(ij)}$
In the notation of \cite{Negahban2012}, the regularization penalty without $\lambda_n$ is expressed as a weighted group norm
\begin{equation}
	{\cal R}(\tilde{\bm \theta}) = \|\tilde{\bm \theta}\|_{\bar{\cal G},2w} 
	  := \sum_{t=1}^{p^2} w_t \| \tilde{\bm \theta}_{Gt} \|_2 
	\label{aeqn705}
\end{equation}
where the index set $\{1,2, \cdots , 2M_n p^2\}$ is partitioned into a set of $N_G=p^2$ disjoint groups $\bar{\cal G} = \{G_1, G_2, \cdots , G_{p^2} \}$ and the subscript $2w$ signifies the weighted group norm. Using this notation, the penalized counterpart to $\tilde{L}_f(\tilde{\bm \Delta})$ of (\ref{admm}) is 
\begin{align}
	\tilde{\cal L}_r(\tilde{\bm \theta}) = & {\cal L}_r(\tilde{\bm \theta})
	        + \lambda_n {\cal R}(\tilde{\bm \theta}) \, .
	\label{aeqn707}
\end{align}

As discussed in \cite[Sec.\ 2.2]{Negahban2012}, w.r.t.\ the usual Euclidean inner product $\langle {\bm u} , {\bm v} \rangle = {\bm u}^\top {\bm v}$ for ${\bm u} , {\bm v} \in \mathbb{R}^{2M_np^2}$ and given any subset $S_{\bar{\cal G}} \subseteq \{1,2, \cdots , N_G \}$ of group indices, define the subspace
\begin{equation}
	{\mathcal M} = \{ \tilde{\bm \theta} \in \mathbb{R}^{2M_np^2} \, | \, 
	      \tilde{\bm \theta}_{Gt} = {\bm 0} \mbox{ for all } t \not\in S_{\bar{\cal G}} \}
	\label{aeqn710}
\end{equation}
and its orthogonal complement
\begin{equation}
	{\mathcal M}^\perp = \{ \tilde{\bm \theta} \in \mathbb{R}^{2M_np^2} \, | \, 
	      \tilde{\bm \theta}_{Gt} = {\bm 0} \mbox{ for all } t \in S_{\bar{\cal G}} \} \, .
	\label{aeqn712}
\end{equation}
The chosen ${\cal R}(\tilde{\bm \theta})$ is decomposable w.r.t.\ $({\mathcal M}, {\mathcal M}^\perp)$ since ${\cal R}(\tilde{\bm \theta}^{(1)}+\tilde{\bm \theta}^{(2)}) = {\cal R}(\tilde{\bm \theta}^{(1)}) + {\cal R}(\tilde{\bm \theta}^{(2)})$ for any $\tilde{\bm \theta}^{(1)} \in {\mathcal M}$ and $\tilde{\bm \theta}^{(2)} \in {\mathcal M}^\perp$ \cite[Sec.\ 2.2, Example 2]{Negahban2012}. 

In order to invoke \cite{Negahban2012}, we need the dual norm ${\cal R}^\circledast$ of regularizer ${\cal R}$ w.r.t.\ the inner product $\langle {\bm u} , {\bm v} \rangle = {\bm u}^\top {\bm v}$ (we use $\circledast$ instead of $\ast$ since $\ast$ has already been used to denote complex conjugation). It is given by \cite[Sec.\ 2.3]{Negahban2012}
\begin{align}
	{\cal R}^\circledast({\bm v}) & = 
	   \sup_{\|{\bm u}\|_{\bar{\cal G},2w} \le 1} \langle {\bm u} , {\bm v} \rangle \; = \;
		\sup_{\|{\bm u}\|_{\bar{\cal G},2w} \le 1} \sum_{i=1}^{2M_n p^2} u_iv_i  \nonumber \\
		& \le \sup_{\|{\bm u}\|_{\bar{\cal G},2w} \le 1} 
		             \sum_{t=1}^{p^2} \|{\bm u}_{Gt}\|_2 \|{\bm v}_{Gt}\|_2 \nonumber \\
		& \le \sup_{\|{\bm u}\|_{\bar{\cal G},2w} \le 1} 
		 \Big( \max_{t \in [p^2]} w_t^{-1} \|{\bm v}_{Gt}\|_2 \Big) 
		    \underbrace{\sum_{t=1}^{p^2} w_t \|{\bm u}_{Gt}\|_2}_{=\|{\bm u}\|_{\bar{\cal G},2w}}  \nonumber \\
	  & \le   \max_{t \in [p^2]} w_t^{-1} \|{\bm v}_{Gt}\|_2 \, . \label{aeqn714}
\end{align}
We also need the subspace compatibility index \cite{Negahban2012}, defined as
\begin{align}
	\Psi({\mathcal M})= & \sup_{{\bm u} \in {\mathcal M} \backslash \{0\}} 
	            {\cal R}({\bm u}) / \|{\bm u }\|_2  \, . \label{aeqn716}
\end{align}
We have ${\cal R}({\bm u}) = \sum_{t=1}^{p^2} w_t \|{\bm u}_{Gt}\|_2 \le 
(\max_{t \in [p^2]} w_t) \sum_{t=1}^{p^2} \|{\bm u}_{Gt}\|_2$. By (\ref{ap2ad13}), $w_t \le 1$ and by the Cauchy-Schwarz inequality, for ${\bm u} \in {\mathcal M}$, $\sum_{t=1}^{p^2} \|{\bm u}_{Gt}\|_2 \le \sqrt{s_n} \, \|{\bm u}\|_2$. Thus, for the lasso and log-sum penalties, $\Psi({\mathcal M}) \le \sqrt{s_n}$. 

We need to establish a restricted strong  convexity condition \cite{Negahban2012} on ${\cal L}_r(\tilde{\bm \theta})$. With $\tilde{\bm \theta}^\diamond$ denoting the true value of $\tilde{\bm \theta}$, let $\tilde{\bm \theta} = \tilde{\bm \theta}^\diamond + \tilde{\bm \gamma}$ with ${\bm \theta}_k = {\bm \theta}_k^\diamond + {\bm \gamma}_k$ (cf.\ (\ref{ap2ad7})). Consider
\begin{align}
	\delta {\cal L}_r(\tilde{\bm \gamma}, \tilde{\bm \theta}^\diamond) 
	:= & {\cal L}_r(\tilde{\bm \theta}^\diamond+\tilde{\bm \gamma}) -
	   {\cal L}_r(\tilde{\bm \theta}^\diamond) 
		-\langle \nabla {\cal L}_r(\tilde{\bm \theta}^\diamond) , \tilde{\bm \gamma} \rangle 
	\label{aeqn720}
\end{align}
where the gradient $\nabla {\cal L}_r(\tilde{\bm \theta}^\diamond)$ at $\tilde{\bm \theta} = \tilde{\bm \theta}^\diamond$ is
\begin{align}
 & \nabla {\cal L}_r(\tilde{\bm \theta}^\diamond) = \begin{bmatrix} 
	(\nabla_1 {\cal L}_r(\tilde{\bm \theta}^\diamond) )^\top & \cdots & 
	  (\nabla_{M_n} {\cal L}_r(\tilde{\bm \theta}^\diamond) )^\top
	   \end{bmatrix}^\top \, , \label{aeqn721} \\
&  \nabla_k {\cal L}_r(\tilde{\bm \theta}^\diamond) 
	  := \frac{\partial{\cal L}_r(\tilde{\bm \theta})}{\partial {\bm \theta}_k} 
		        \Big|_{\tilde{\bm \theta} = \tilde{\bm \theta}^\diamond} 
						=  2 {\cal H}_k {\bm \theta}_k^\diamond- 2\, {\bm b}_k  \, .
	\label{aeqn722}
\end{align}
Noting that ${\cal H}_k = {\cal H}_k^\top$,  (\ref{aeqn720}) simplifies to 
\begin{align}
	\delta {\cal L}_r(\tilde{\bm \gamma}, \tilde{\bm \theta}^\diamond) & 
	= \sum_{k=1}^{M_n} {\bm \gamma}_k^\top  {\cal H}_k {\bm \gamma}_k  \, ,
	\label{aeqn724}
\end{align}
which may be rewritten as 
\begin{align}
	\delta {\cal L}_r(\tilde{\bm \gamma}, \tilde{\bm \theta}^\diamond) & 
	= \sum_{k=1}^{M_n}  \Big[ {\bm \gamma}_k^\top  {\cal H}_k^\diamond {\bm \gamma}_k
	   + {\bm \gamma}_k^\top \big( {\cal H}_k-{\cal H}_k^\diamond \big) {\bm \gamma}_k \Big]  \, .
	\label{aeqn726}
\end{align}

Under the sparsity assumption (\ref{heqn200}), $\tilde{\bm \theta}^\diamond = \tilde{\bm \theta}^\diamond_{\mathcal M}$, hence, $\tilde{\bm \theta}^\diamond_{{\mathcal M}^\perp} = {\bm 0}$, where  $\tilde{\bm \theta}_{\mathcal M}$ and $\tilde{\bm \theta}_{{\mathcal M}^\perp}$ denote projection of $\tilde{\bm \theta}$ on subspaces ${\mathcal M}$ and ${\mathcal M}^\perp$, respectively. Similar to $\hat{\tilde{\bm \Delta}} = \arg \min_{\tilde{\bm \Delta}} \tilde{L}_f(\tilde{\bm \Delta})$, suppose
\begin{equation}
  \hat{\tilde{\bm \theta}} = \arg\min_{\tilde {\bm \theta}} 
	      \big\{ {\cal L}_r(\tilde{\bm \theta})
	        + \lambda_n {\cal R}(\tilde{\bm \theta}) \big\} \, , \label{aeqn730}
\end{equation}
and we consider (\ref{aeqn720}) and (\ref{aeqn724}) with $\hat{\tilde{\bm \theta}} = \tilde{\bm \theta}^\diamond + \tilde{\bm \gamma}$. Then
\begin{align}
  \hat{\tilde{\bm \theta}} & - \tilde{\bm \theta}^\diamond 
	 = \hat{\tilde{\bm \theta}}_{\mathcal M} - \tilde{\bm \theta}^\diamond 
	   + \hat{\tilde{\bm \theta}}_{{\mathcal M}^\perp} 
	= \tilde{\bm \gamma}_{\mathcal M} + \tilde{\bm \gamma}_{{\mathcal M}^\perp}  \, . \label{aeqn732}
\end{align}
By \cite[Lemma 1]{Negahban2012},
\begin{align}
  {\cal R}( \tilde{\bm \gamma}_{{\mathcal M}^\perp} ) \; \le \; & 
	 3 {\cal R}( \tilde{\bm \gamma}_{{\mathcal M}} ) 
	  + 4  {\cal R}(  \tilde{\bm \theta}^\diamond_{{\mathcal M}^\perp}) \, , \label{aeqn734}
\end{align}
if we pick
\begin{align}
  \lambda_n \ge & 2 ~ {\cal R}^\circledast(\nabla {\cal L}_r(\tilde{\bm \theta}^\diamond))  \, . \label{aeqn736}
\end{align}
Since in our case $\tilde{\bm \theta}^\diamond_{{\mathcal M}^\perp} = {\bm 0}$, we have ${\cal R}(  \tilde{\bm \theta}^\diamond_{{\mathcal M}^\perp})=0$. 

We now turn to bounding ${\cal R}^\circledast(\nabla {\cal L}_r(\tilde{\bm \theta}^\diamond))$. First we need several auxiliary results. Define  
\begin{align}
  {\bm \Delta}_{xk} := & \hat{\bm S}_{xk} -{\bm S}_{xk}^\diamond \, , \quad
	{\bm \Delta}_{yk} :=  \hat{\bm S}_{yk} -{\bm S}_{yk}^\diamond \, ,  \label{aeqn400} \\
	{\bm \Delta}_{yxk} := & \hat{\bm S}_{yk}^\ast \otimes \hat{\bm S}_{xk} 
	  - ({\bm S}_{yk}^\diamond)^\ast \otimes {\bm S}_{xk}^\diamond  \, , \label{aeqn401} \\ 
		\bar{\delta}_x = & \max_{k \in M_n} \|{\bm \Delta}_{xk}\|_\infty \, , \quad
		\bar{\delta}_y =  \max_{k \in M_n} \|{\bm \Delta}_{yk}\|_\infty \, , \label{aeqn402} \\
		\bar{\delta} \ge & \max\{ \bar{\delta}_x, \bar{\delta}_y \} \,.  \label{aeqn403}	
\end{align}
{\it Lemma 1}.  Under (\ref{aeqn400})-(\ref{aeqn403}) and with $B_{xy}$ as in (\ref{heqn320}), we have
\begin{align}
  \| {\bm \Delta}_{yxk} \|_\infty &  \le  \bar{\delta}^2 + 2 \, B_{xy} \, \bar{\delta} =: \bar{B} \, .
    \label{aeqn405}	
\end{align}
{\it Poof}. We can rewrite ${\bm \Delta}_{yxk}$ as
\begin{align}
	{\bm \Delta}_{yxk} = & {\bm \Delta}_{yk}^\ast \otimes {\bm \Delta}_{xk} 
	  + ({\bm S}_{yk}^\diamond)^\ast \otimes {\bm \Delta}_{xk} 
		+ {\bm \Delta}_{yk}^\ast \otimes {\bm S}_{xk}^\diamond  . \label{aeqn420}
\end{align}
Therefore
\begin{align}
	& \| {\bm \Delta}_{yxk} \|_\infty \le    \| {\bm \Delta}_{yk} \|_\infty \, \| {\bm \Delta}_{xk} \|_\infty
	  + \| ({\bm S}_{yk}^\diamond \|_\infty \, \| {\bm \Delta}_{xk} \|_\infty  \nonumber \\
		& \quad\quad\quad + \| {\bm \Delta}_{yk} \|_\infty \, \| {\bm S}_{xk}^\diamond \|_\infty \nonumber \\
	&	\; \le  \bar{\delta}_y  \bar{\delta}_x + B_{xy}  \bar{\delta}_x + \bar{\delta}_y   B_{xy}
		\le \bar{\delta}^2 + 2 B_{xy}  \bar{\delta} \, . \quad \blacksquare 
\end{align}

Using the notation $G_t$ for the group $t$ corresponding to the edge $\{i,j\}$, as in (\ref{ap2ad10}), let $(\nabla {\cal L}_r(\tilde{\bm \theta}^\diamond))_{Gt} \in \mathbb{R}^{2M}$ denote the corresponding  entries of the gradient. By (\ref{aeqn721})-(\ref{aeqn722}), we have 
\begin{align}
 & (\nabla {\cal L}_r(\tilde{\bm \theta}^\diamond))_{Gt} = \begin{bmatrix} 
	(\nabla_1 {\cal L}_r(\tilde{\bm \theta}^\diamond) )_{Gt}^\top & \cdots & 
	  (\nabla_{M_n} {\cal L}_r(\tilde{\bm \theta}^\diamond) )_{Gt}^\top
	   \end{bmatrix}^\top \, , \label{aeqn1721} \\
&  (\nabla_k {\cal L}_r(\tilde{\bm \theta}^\diamond))_{Gt} 
	 =  (2 {\cal H}_k {\bm \theta}_k^\diamond- 2\, {\bm b}_k)_{Gt} 
	   \in \mathbb{R}^{2}\, . \label{aeqn1722}  
\end{align}
At the true values ${\cal H}_k = {\cal H}_k^\diamond$ and ${\bm b}_k = {\bm b}_k^\diamond$, 
\begin{align}
   &   \nabla_k {\cal L}_r(\tilde{\bm \theta}^\diamond)\Big|_
			 {{\cal H}_k={\cal H}_k^\diamond, {\bm b}_k={\bm b}_k^\diamond} = {\bm 0} 
			 =  2{\cal H}_k^\diamond {\bm \theta}_k^\diamond- 2\, {\bm b}_k^\diamond
\end{align} 
(cf.\ (\ref{13eqn15})-(\ref{15eqn15})) where
\begin{align}
	{\cal H}_k^\diamond & := \frac{1}{2} {\bm T}_{rc}^H  \bar{\cal H}_k^\diamond {\bm T}_{rc}
						 \, ,  \quad
		{\bm b}_k^\diamond  := \frac{1}{2} {\bm T}_{rc}^H \bar{\bm b}_k^\diamond 	 \label{ap2ad82} \\
		\bar{\cal H}_k^\diamond & := 
		 \begin{bmatrix} ({\bm S}_{yk}^\diamond)^\ast \otimes {\bm S}_{xk}^\diamond  & {\bm 0} \\
	           {\bm 0} & {\bm S}_{yk}^\diamond \otimes ({\bm S}_{xk}^\diamond)^\ast \end{bmatrix} \, ,
		\label{ap2ad84} \\
		\bar{\bm b}_k^\diamond & = \begin{bmatrix} \mbox{vec}({\bm S}_{xk}^\diamond-{\bm S}_{yk}^\diamond) \\ 
					 \mbox{vec}(({\bm S}_{xk}^\diamond-{\bm S}_{yk}^\diamond)^\ast) 
					 \end{bmatrix} \, . \label{ap2ad86}
\end{align}
Therefore, we may rewrite (\ref{aeqn1722}) as
\begin{align}
&  (\nabla_k {\cal L}_r(\tilde{\bm \theta}^\diamond))_{Gt} 
	 =  ( 2 ({\cal H}_k -{\cal H}_k^\diamond) {\bm \theta}_k^\diamond
	  - 2\,( {\bm b}_k- {\bm b}_k^\diamond) )_{Gt}   \nonumber \\
&  = \sum_{q=1}^{p^2} \Big[ 2 ({\cal H}_k-{\cal H}_k^\diamond)_{Gt,Gq} ({\bm \theta}_k^\diamond)_{Gq} \Big]
          - 2\, ({\bm b}_k - {\bm b}_k^\diamond)_{Gt}  
	\label{aeqn1723}
\end{align}
where $G_q$ represents group $q$ corresponding to some edge $\{\ell, m\}$,  $(i,j) \leftrightarrow t = (i-1)p+j$ and $(\ell, m) \leftrightarrow q = (\ell-1)p+m$. 

{\it Lemma 2}.  Under the conditions of Lemma 1
\begin{align}
  \| (\nabla_k {\cal L}_r(\tilde{\bm \theta}^\diamond))_{Gt} \|_2 &  
	  \le  \sum_{q=1}^{p^2} 2  ~ \bar{B}
			 ~ \| ({\bm \theta}_k^\diamond)_{Gq} \|_2  + 4 \bar{\delta} \, .
    \label{aeqn4050}	
\end{align}
{\it Proof}. With $(i,j) \leftrightarrow t = (i-1)p+j$ and $(\ell, m) \leftrightarrow q = (\ell-1)p+m$, we have
\begin{align}
		& \big( \bar{\cal H}_k -\bar{\cal H}_k^\diamond \big)_{Gt,Gq}  = 
		 \begin{bmatrix} a_k  & 0 \\
	           0 & a_k^\ast \end{bmatrix} \, , \label{ap2ad100} \\
		&	a_k   := [\hat{\bm S}_{yk}^\ast]_{jm} [\hat{\bm S}_{xk}]_{i \ell}
				   -[({\bm S}_{yk}^\diamond)^\ast]_{jm} [{\bm S}_{xk}^\diamond]_{i \ell} \, ,
		\label{ap2ad101} \\
	&	( \bar{\bm b}_k- \bar{\bm b}_k^\diamond) )_{Gt}  = \begin{bmatrix} 
		    [{\bm \Delta}_{xk} - {\bm \Delta}_{yk}]_{ij} \\ 
					 [{\bm \Delta}_{xk} - {\bm \Delta}_{yk}]_{ij}^\ast
					 \end{bmatrix} \, , \label{ap2ad102}
\end{align}
where (\ref{ap2ad100})-(\ref{ap2ad101}) follow from $\mbox{vec}({\bm S}_{xk} {\bm \Delta}_k {\bm S}_{yk}) = ({\bm S}_{yk}^\top \otimes {\bm S}_{xk}) \mbox{vec}({\bm \Delta}_k)$ and ${\bm S}_{yk}^\top = {\bm S}_{yk}^\ast$. Using Lemma 1, $\|\tilde{\bm T}_{rc}\| = \sqrt{2}$ and $\|(\bar{\cal H}_k - \bar{\cal H}_k^\diamond)_{Gt,Gq}\| \le \| {\bm \Delta}_{yxk} \|_\infty$, we have
\begin{align}
  & \|2 ({\cal H}_k-{\cal H}_k^\diamond)_{Gt,Gq}\| = 
	     \| \tilde{\bm T}_{rc}^H (\bar{\cal H}_k - \bar{\cal H}_k^\diamond)_{Gt,Gq} 
			           \tilde{\bm T}_{rc} \|  \nonumber \\
	& \quad \le \| \tilde{\bm T}_{rc}^H \|  ~ \|(\bar{\cal H}_k - \bar{\cal H}_k^\diamond)_{Gt,Gq}\|
	     ~ \| \tilde{\bm T}_{rc} \| \le 2  ~ \bar{B} \, . \label{ap2ad105}
\end{align}
By (\ref{aeqn400}), (\ref{aeqn402}), (\ref{aeqn403}) and (\ref{ap2ad102})
\begin{align}
  \| 2\, ({\bm b}_k - {\bm b}_k^\diamond)_{Gt} \|_2 & = 
	     \| \tilde{\bm T}_{rc}^H ( \bar{\bm b}_k- \bar{\bm b}_k^\diamond) )_{Gt} \|  \nonumber \\
	& 	\le \| \tilde{\bm T}_{rc}^H \|  ~ \|( \bar{\bm b}_k- \bar{\bm b}_k^\diamond) )_{Gt}\|_2
	      \le 4 \bar{\delta} \, . \label{ap2ad107}
\end{align}
By (\ref{aeqn1723}), (\ref{ap2ad105}) and (\ref{ap2ad107}) we have (\ref{aeqn4050}). $\quad \blacksquare$

{\it Lemma 3}.  Under the conditions of Lemma 1 if $\bar{\delta} \le B_{xy}$,
\begin{align}
  & {\cal R}^\circledast(\nabla {\cal L}_r(\tilde{\bm \theta}^\diamond))
	   \le B_{\rm init} \sqrt{M_ n} ~ \big( 6 \, B_{xy} B_d s_n + 4 \big) ~ \bar{\delta}
   \label{aeqn4060}	
\end{align}
where $B_d$ and $B_{\rm init}$ are given by (\ref{heqn321}) and (\ref{heqn327}), respectively.  \\
{\it Proof}. By Lemma 2 and (\ref{aeqn1721}), 
\begin{align}
 & \| ( \nabla {\cal L}_r(\tilde{\bm \theta}^\diamond) )_{Gt} \|_2  =
    \sqrt{ \sum_{k=1}^{M_n} \| (\nabla_k {\cal L}_r(\tilde{\bm \theta}^\diamond))_{Gt} \|_2^2 } \nonumber \\
	& \; \le \sqrt{M_ n} ~  \max_{k \in [M_n]} ~ 
	       \| (\nabla_k {\cal L}_r(\tilde{\bm \theta}^\diamond))_{Gt} \|_2  \nonumber \\
	& \; \le  \sqrt{M_ n} \big[ 2 ~ \bar{B} \max_{k \in [M_n]} 
	   \big( \sum_{q=1}^{p^2} \| ({\bm \theta}_k^\diamond)_{Gq} \|_2 \big) + 4  \bar{\delta} \big] \, .
    \label{aeqn4070}	
\end{align}
Observe that $\sum_{q=1}^{p^2} \| ({\bm \theta}_k^\diamond)_{Gq} \|_2 \le s_n ~ \max_{q \in [p^2]} \| ({\bm \theta}_k^\diamond)_{Gq} \|_2$ since at most $s_n$ edges are connected in the true graph. For group $q$ with $(\ell, m) \leftrightarrow q = (\ell-1)p+m$, $\| ({\bm \theta}_k^\diamond)_{Gq} \|_2 = | [{\bm \Delta}_k^\diamond]_{\ell m} | \le B_d$ for $k \in [M_n]$. Therefore, 
\begin{align}
 & \| ( \nabla {\cal L}_r(\tilde{\bm \theta}^\diamond) )_{Gt} \|_2  
  \le  \sqrt{M_ n} \Big[ 2 ~ \bar{B} s_n B_d + 4  \bar{\delta} \Big] \, .
    \label{aeqn4075}	
\end{align}
By (\ref{aeqn714}) and (\ref{aeqn4075})
\begin{align}
  & {\cal R}^\circledast(\nabla {\cal L}_r(\tilde{\bm \theta}^\diamond))
	   \le \max_{t \in [p^2]} w_t^{-1} \| ( \nabla {\cal L}_r(\tilde{\bm \theta}^\diamond) )_{Gt} \|_2 \nonumber \\
	& \quad 
	   \le (\max_{t \in [p^2]} w_t^{-1}) ~ (\max_{t \in [p^2]} 
		  \| ( \nabla {\cal L}_r(\tilde{\bm \theta}^\diamond) )_{Gt} \|_2) \nonumber \\
	& \quad \overset{\bar{\delta} \le B_{xy}}{\le}  B_{\rm init} 
	    \sqrt{M_ n} ~ \big( 6 \, B_{xy} B_d s_n + 4 \big) ~ \bar{\delta}
   \label{aeqn4080}	
\end{align}
where, for the log-sum penalty we used $\max_{t \in [p^2]} w_t^{-1} = 1 + \max_{t \in [p^2]} \|\tilde{\bm \theta}_{Gt} \| /\epsilon  = 1+ \max_{i,j \in [p]} \| \bar{\tilde{\bm \Delta}}^{(ij)} \| / \epsilon =: B_{\rm init}$, and  $\bar{\delta} \le B_{xy}$ results in $\bar{B} = \bar{\delta}^2 + 2 \, B_{xy} \le 3  B_{xy} \bar{\delta}$. $\quad \blacksquare$

{\it Lemma 4}.  Under the conditions of Lemmas 1 and 3, if $\lambda_n \ge 
2 {\cal R}^\circledast(\nabla {\cal L}_r(\tilde{\bm \theta}^\diamond))$, 
\begin{align}
   \delta {\cal L}_r(\tilde{\bm \gamma}, \tilde{\bm \theta}^\diamond)  
	\ge \kappa_{\cal L} \, \| \tilde{\bm \gamma} \|_2^2 \, , 
	        \label{aeqn749}
\end{align}
where $\kappa_{\cal L} = \phi_{\min}^\diamond - 192 s_n M_n B_{\rm init}^2 B_{xy} \bar{\delta}  $.  \\
{\it Proof}. Consider (\ref{aeqn726}). By (\ref{ap2ad8a}) we have
\begin{align}
 & \sum_{k=1}^{M_n}   {\bm \gamma}_k^\top  {\cal H}_k^\diamond {\bm \gamma}_k
  = \sum_{k=1}^{M_n}  \frac{1}{2} ({\bm T}_{rc} {\bm \gamma}_k)^H  
	        \bar{\cal H}_k^\diamond ({\bm T}_{rc} {\bm \gamma}_k)  \nonumber \\
& \quad \ge \sum_{k=1}^{M_n}  \frac{1}{2} \phi_{\min}(\bar{\cal H}_k^\diamond) 
           \|{\bm T}_{rc} {\bm \gamma}_k\|_2^2 \nonumber \\
& \quad
   = \sum_{k=1}^{M_n}  \phi_{\min}(\bar{\cal H}_k^\diamond) \|{\bm \gamma}_k\|_2^2
	  \; \mbox{ since } \; {\bm T}_{rc}^H {\bm T}_{rc} = 2 {\bm I}_{p^2} \, .
	\label{aeqn40100}
\end{align}
Now $\phi_{\min}(\bar{\cal H}_k^\diamond) = \phi_{\min}({\bm S}_{yk}^\diamond) \phi_{\min}({\bm S}_{xk}^\diamond) \ge \phi_{\min}^\diamond$, implying
\begin{align}
 & \sum_{k=1}^{M_n}   {\bm \gamma}_k^\top  {\cal H}_k^\diamond {\bm \gamma}_k
  \ge \phi_{\min}^\diamond \sum_{k=1}^{M_n}  \|{\bm \gamma}_k\|_2^2  
	 = \phi_{\min}^\diamond  \|\tilde{\bm \gamma}\|_2^2 \, .
	\label{aeqn40102}
\end{align}
Define
\begin{align}
 \check{\cal H} & := \mbox{block-diag} \big\{ {\cal H}_1 \, , \; \cdots , \; {\cal H}_{M_n} \big\} \, ,
	\label{aeqn40110} \\
	\check{\cal H}^\diamond & := \mbox{block-diag} \big\{ {\cal H}_1^\diamond \, , \; \cdots , 
	           \; {\cal H}_{M_n}^\diamond \big\} \, .  \label{aeqn40112}
\end{align}
We have $\| \check{\cal H}-\check{\cal H}^\diamond \|_\infty  = \max_{k \in [M_n]} 
\| {\cal H}_k-{\cal H}_k^\diamond \|_\infty $.  Using the facts $\|{\bm A} {\bm B}\|_\infty \le \|{\bm A}\|_\infty ~ \| {\bm B}^\top \|_{1,\infty}$ and $\|{\bm A} {\bm B}\|_\infty \le \|{\bm B}\|_\infty ~ \| {\bm A} \|_{1,\infty}$, and Lemma 1, we have
\begin{align}
  & \| {\cal H}_k-{\cal H}_k^\diamond \|_\infty  = \| \frac{1}{2} {\bm T}_{rc}^H 
	       (\bar{\cal H}_k - \bar{\cal H}_k^\diamond) {\bm T}_{rc} \|_\infty  \nonumber \\
	&  \le \frac{1}{2} \| {\bm T}_{rc}^\top \|_{1,\infty}^2 
	        \| \bar{\cal H}_k - \bar{\cal H}_k^\diamond \|_\infty
					\le \frac{4}{2}  \| {\bm \Delta}_{yxk} \|_\infty \le 2 \bar{B} \, . \label{aeqn40115}
\end{align}
By (\ref{aeqn40110})-(\ref{aeqn40115}),
\begin{align}
 & | \sum_{k=1}^{M_n}   {\bm \gamma}_k^\top  ({\cal H}_k-{\cal H}_k^\diamond) {\bm \gamma}_k |
  = | \tilde{\bm \gamma}^\top \big( \check{\cal H} - \check{\cal H}^\diamond \big)  \tilde{\bm \gamma} | \nonumber \\
& \quad \le \sum_{\ell=1}^{2p^2 M_n} \sum_{m=1}^{2p^2 M_n} | \tilde{\gamma}_\ell 
    \big[ \check{\cal H} - \check{\cal H}^\diamond \big]_{\ell m}  \tilde{\gamma}_m | \nonumber \\
& \quad \le \| \check{\cal H}-\check{\cal H}^\diamond \|_\infty 
   \Big(\sum_{m=1}^{2p^2 M_n} |\tilde{\gamma}_m | \Big)^2  =: A \, .
	\label{aeqn40117}
\end{align}
As in (\ref{ap2ad7}), $\tilde{\bm \gamma} = \big[ {\bm \gamma}_1^\top , \; \cdots , \; {\bm \gamma}^\top \big]^\top$. Expressing in terms of group $G_t$ and using the Cauchy-Schwarz inequality, we have
\begin{align}
 & \sum_{m=1}^{2p^2 M_n} |\tilde{\gamma}_m | = \sum_{t=1}^{p^2} 
         \Big( \sum_{k=1}^{M_n} \big[  |\gamma_{kt} | +|\gamma_{k(t+p^2)} |  \big] \Big)  \nonumber \\
	& 	\le \sum_{t=1}^{p^2} \sqrt{2 M_n} \, \|\tilde{\bm \gamma}_{Gt} \|_2   
	\le  \sqrt{2 M_n}  \, \big( \max_{t \in [p^2]} w_t^{-1} \big) 
	      \sum_{t=1}^{p^2} w_t \|\tilde{\bm \gamma}_{Gt} \|_2  \nonumber \\
	& \quad 
				  \le \sqrt{2 M_n} \,  B_{\rm init} \|\tilde{\bm \gamma}\|_{\bar{\cal G},2w}  \, .
	\label{aeqn40120}
\end{align}
Thus by (\ref{aeqn40115}), (\ref{aeqn40117}) and (\ref{aeqn40120}), if $\bar{\delta} \le B_{xy}$, 
\begin{align}
 A & \le  12 B_{\rm init}^2 M_n B_{xy} ~\bar{\delta} ~\|\tilde{\bm \gamma}\|_{\bar{\cal G},2w}^2 \, .
	\label{aeqn40122}
\end{align}
By (\ref{aeqn734}) and (\ref{aeqn736}) we have
\begin{align}
 & \|\tilde{\bm \gamma}\|_{\bar{\cal G},2w}^2 
  = \| \tilde{\bm \gamma}_{\mathcal M} + \tilde{\bm \gamma}_{{\mathcal M}^\perp} \|_{\bar{\cal G},2w}^2 \nonumber \\
	& \quad
	= (\| \tilde{\bm \gamma}_{\mathcal M} \|_{\bar{\cal G},2w} 
	  + \|\tilde{\bm \gamma}_{{\mathcal M}^\perp} \|_{\bar{\cal G},2w})^2 \nonumber \\
	& \quad \overset{(\ref{aeqn734})}{\le} 16 \, \| \tilde{\bm \gamma}_{\mathcal M} \|_{\bar{\cal G},2w}^2
	   \overset{(\ref{aeqn716})}{\le} 16 s_n \| \tilde{\bm \gamma}_{\mathcal M} \|_2^2 
		 \le 16 s_n \| \tilde{\bm \gamma} \|_2^2
  \, . \label{aeqn40125}
\end{align}
Using (\ref{aeqn726}), (\ref{aeqn40102}), (\ref{aeqn40117}), (\ref{aeqn40122}) and (\ref{aeqn40125}), we have
\begin{align}
 & \delta {\cal L}_r(\tilde{\bm \gamma}, \tilde{\bm \theta}^\diamond)  \ge 
   \big( \phi_{\min}^\diamond - 192 s_n M_n B_{\rm init}^2 B_{xy} \bar{\delta} \big) \| \tilde{\bm \gamma} \|_2^2 
	= \kappa_{\cal L} \, \| \tilde{\bm \gamma} \|_2^2
  \, , \nonumber 
\end{align}
proving the desired result. $\quad \blacksquare$

Using \cite[Lemma 1]{Tugnait2022} we have Lemma 5. \\
{\it Lemma 5}. Let $\sigma_{xy}$,  ${C}_0$ and $N_1$ be as in (\ref{heqn323}), (\ref{heqn324}) and (\ref{heqn325}), respectively. Define 
\begin{align*} 
 &  {\mathcal A} =   \max_{k \in [M_n], \, q, \ell \in [p_n]} 
	      \big\{ \big| [ \hat{\bm S}_{xk} - {\bm S}_{xk}^\diamond ]_{q \ell} \big|  , 
	 \big| [ \hat{\bm S}_{yk} - {\bm S}_{yk}^\diamond ]_{q \ell} \big| \big\}  . 
\end{align*}
Then for any $\tau > 2$ and sample size $n > N_1$,
\begin{align}  
   P & \Big(  {\mathcal A}  > {C}_0 \sqrt{\ln(p_n)/K_n} \Big) \le 2/p_n^{\tau -2} \, . \label{aeqn315}
\end{align}
{\it Proof}. By \cite[Lemma 1]{Tugnait2022}, 
\begin{align}  
   P \Big(  \max_{k , q, \ell } &\big| [ \hat{\bm S}_{xk} - {\bm S}_{xk}^\diamond ]_{q \ell}\big|
	    > C_{0x} \sqrt{\frac{\ln(p_n )}{K_n}} \Big)  \le \frac{1}{p_n ^{\tau -2}} \label{naeq58b} \\
	  P \Big(  \max_{k , q, \ell } &\big| [ \hat{\bm S}_{yk} - {\bm S}_{yk}^\diamond ]_{q \ell} \big|
	    > C_{0y} \sqrt{\frac{\ln(p_n )}{K_n}} \Big)  \le \frac{1}{p_n ^{\tau -2}} \label{naeq58c} 
\end{align}
for any $\tau > 2$ and sample size $n > N_1$ where $C_{0x} = 80 \,\max_{\ell, f} ( [{\bm S}_x^\diamond(f)]_{\ell \ell}) 
	    \sqrt{ N_1 / \ln ( p_n )  }$ and 
$C_{0y} = 80 \,\max_{\ell, f} ( [{\bm S}_y^\diamond(f)]_{\ell \ell}) 
	    \sqrt{ N_1 / \ln ( p_n )  }$. 
Using the union bound,
\begin{align}  
   P & \Big(  {\mathcal A}  > C_0 \sqrt{\ln(p_n)/ K_n} \Big) \nonumber \\
	  \le & P \Big(  \max_{k , q, \ell } \big| [ \hat{\bm S}_{xk} - {\bm S}_{xk}^\diamond ]_{q \ell}\big|
	    > C_0 \sqrt{\ln(p_n)/ K_n} ~ \Big) \nonumber \\ 
			& \; + P \Big(  \max_{k , q, \ell } \big| [ \hat{\bm S}_{yk} - {\bm S}_{yk}^\diamond ]_{q \ell}\big|
	    > C_0 \sqrt{\ln(p_n)/ K_n} ~ \Big)  \nonumber \\
	  \le & 2/p_n^{\tau -2}  \label{aeqn316}
\end{align}
since $C_0 \ge C_{0x}$ and $C_0 \ge C_{0y}$. $\quad \blacksquare$

We are now ready to prove Theorem 1. \\
{\it Proof of Theorem 1}. First choose $\bar{\delta}$ to make $\kappa_{\cal L} > 0$ in Lemma 4. Suppose we take $192 s_n M_n B_{\rm init}^2 B_{xy} \bar{\delta} \le \phi_{\min}^\diamond/4$. Then $\kappa_{\cal L} \ge 3 \phi_{\min}^\diamond/4$. Now pick 
\begin{align}
 \bar{\delta} = & C_0 \sqrt{ln(p_n)/K_n} \; \le \; \min \big\{ B_{xy}, 
  \frac{\phi_{\min}^\diamond}{768 s_n M_n B_{\rm init}^2 B_{xy}} \big\}
  \, ,  \label{aeqn770}
\end{align}
leading to $192 s_n M_n B_{\rm init}^2 B_{xy} \bar{\delta} \le  \phi_{\min}^\diamond/4$. These upper bounds can be ensured by picking appropriate lower bounds to sample size $n$ and invoking Lemma 5. The choice of $n$ specified in (\ref{eqn452}) satisfies (\ref{aeqn770}) with probability $> 1- 2/p_n^{\tau -2}$. Using $\bar{\delta} = C_0 \sqrt{ln(p_n)/K_n} \le B_{xy}$, the lower bound on $\lambda_n$ given in (\ref{eqn450}) satisfies (\ref{aeqn736}) with ${\cal R}^\circledast(\nabla {\cal L}_r(\tilde{\bm \theta}^\diamond))$ as in Lemma 3. By \cite[Theorem 1]{Negahban2012}, $\hat{\tilde{\bm \theta}}$ given by (\ref{aeqn730}) satisfies 
\begin{align}
 \|\hat{\tilde{\bm \theta}}- \tilde{\bm \theta}^\diamond \|_2 \, \le \, & 
 \frac{3 \lambda_n}{\kappa_{\cal L}} \Psi({\mathcal M})
  \, .  \label{aeqn772}
\end{align}
The left-side of (\ref{aeqn772}) equals $\| \hat{\tilde{\bm \Delta}} - \tilde{\bm \Delta}^\diamond \|_F$ while the right-side of (\ref{aeqn772}) equals the last term of (\ref{eqn454}) using $\Psi({\mathcal M}) \le \sqrt{s_n}$, $\kappa_{\cal L} \ge 3 \phi_{\min}^\diamond/4$. This proves Theorem 1. $\quad \blacksquare$

We now turn to the proof of Theorem 2. \\
{\it Proof of Theorem 2}. We have $\| \hat{\tilde{\bm \Delta}}^{(ij)} - (\tilde{\bm \Delta}^\diamond)^{(ij)} \| \le \| \hat{\tilde{\bm \Delta}} - \tilde{\bm \Delta}^\diamond \|_F \le \bar{\sigma}_n$ w.h.p. For the edge $\{ i,j \} \in \tilde{\cal E}_\Delta^\diamond$, we have
\begin{align}  
		   \| \hat{\tilde{\bm \Delta}}^{(ij)} \| = & 
			    \| (\tilde{\bm \Delta}^\diamond)^{(ij)} + \hat{\tilde{\bm \Delta}}^{(ij)} 
					   - (\tilde{\bm \Delta}^\diamond)^{(ij)} \|  \nonumber \\
		 \ge & \| (\tilde{\bm \Delta}^\diamond)^{(ij)} \|
		       - \| \hat{\tilde{\bm \Delta}}^{(ij)}  - (\tilde{\bm \Delta}^\diamond)^{(ij)} \| \nonumber \\
		 \ge & \nu - \bar{\sigma}_n \ge 0.6 \, \nu \;\; \mbox{ for } \;\; n \ge N_4 \nonumber \\
		 > & \gamma_n \, .
		\label{naeq8397} 
\end{align}
Thus, $\tilde{\cal E}_\Delta^\diamond \subseteq \hat{\cal E}_\Delta$. Now consider the set complements $(\tilde{\cal E}_\Delta^\diamond)^c$ and $\hat{\cal E}_\Delta^c$. For the edge $\{ i.j \} \in (\tilde{\cal E}_\Delta^\diamond)^c$, $\| (\tilde{\bm \Delta}^\diamond)^{(ij)} \| = 0$. For $n \ge N_4$, w.h.p.\ we have
\begin{align}  
		   \| \hat{\tilde{\bm \Delta}}^{(ij)} \| \le & 
			 \| (\tilde{\bm \Delta}^\diamond)^{(ij)} \| 
			     + \| \hat{\tilde{\bm \Delta}}^{(ij)} \hat{\tilde{\bm \Delta}}^{(ij)}  
					 - (\tilde{\bm \Delta}^\diamond)^{(ij)} \| \nonumber \\
		 \le & 0 + \bar{\sigma}_n \le 0.4 \, \nu <  \gamma_n \, ,
		\label{naeq8399} 
\end{align}
implying that $\{ i,j \} \in \hat{\cal E}_\Delta^c$. Thus, $(\tilde{\cal E}_\Delta^\diamond)^c \subseteq \hat{\cal E}^c$, hence $\hat{\cal E}_\Delta \subseteq \tilde{\cal E}_\Delta^\diamond$, establishing $\hat{\cal E}_\Delta = \tilde{\cal E}_\Delta^\diamond$. $\quad \blacksquare$

\bibliographystyle{unsrt}

\end{document}